\documentclass[10pt,twocolumn,letterpaper]{article}

\usepackage{iccv}
\usepackage{times}
\usepackage{epsfig}
\usepackage{graphicx}
\usepackage{amsmath}
\usepackage{amssymb}

\usepackage{booktabs}
\usepackage{adjustbox}
\usepackage{xcolor}
\usepackage{multirow}

\usepackage{algorithm}
\usepackage{algorithmic}
\usepackage{pifont}%
\usepackage{subfigure}

\usepackage{dsfont}
\usepackage{minitoc}
\usepackage{placeins}

\newcommand{\RomanNumeral}[1]
    {\romannumeral #1}
\newcommand{\cmark}{\textcolor{black}{\ding{51}}}%
\newcommand{\xmark}{\textcolor{black}{\ding{55}}}%

\newcommand{\refsupp}[1]{{\bf\color{red} #1}} 
\usepackage[pagebackref=true,breaklinks=true,letterpaper=true,colorlinks,bookmarks=false]{hyperref}

\iccvfinalcopy %

\def\iccvPaperID{12229} %

\ificcvfinal\fi

\begin{document}

\title{Strata-NeRF : Neural Radiance Fields for Stratified Scenes}

\author{
Ankit Dhiman\textsuperscript{1,2} \quad
R Srinath\textsuperscript{1} \quad
Harsh Rangwani\textsuperscript{1} \quad
Rishubh Parihar\textsuperscript{1} \\
Lokesh R Boregowda\textsuperscript{2} \quad 
Srinath Sridhar\textsuperscript{3} \quad
R Venkatesh Babu\textsuperscript{1} \\
\textsuperscript{1}Vision and AI Lab, IISc Bangalore \quad
\textsuperscript{2}Samsung R \& D Institute India - Bangalore \quad
\textsuperscript{3}Brown University \\
}

\maketitle
\ificcvfinal\thispagestyle{empty}\fi

\begin{abstract}

        Neural Radiance Field (NeRF) approaches learn the underlying 3D representation of a scene and generate photo-realistic novel views with high fidelity.  However, most proposed settings concentrate on modelling a single object or a single level of a scene.  However, in the real world, we may capture a scene at multiple levels, resulting in a layered capture. For example, tourists usually capture a monument's exterior structure before capturing the inner structure. Modelling such scenes in 3D with seamless switching between levels can drastically improve
        immersive experiences.
        However, most existing techniques struggle in modelling such scenes. We propose \emph{Strata-NeRF}, a single neural radiance field that 
        implicitly captures a scene with multiple levels.
        \emph{Strata-NeRF} achieves this by conditioning the NeRFs on Vector Quantized (VQ) latent representations which allow sudden changes in scene structure.
        We evaluate the effectiveness of our 
        approach in multi-layered synthetic dataset comprising diverse scenes and then further validate its generalization on the real-world RealEstate10K dataset.
        We find that \emph{Strata-NeRF} effectively captures stratified scenes, minimizes artifacts, and synthesizes high-fidelity views compared to existing approaches. \url{https://ankitatiisc.github.io/Strata-NeRF/}
\end{abstract}

\section{Introduction}
\label{sec:intro}

\begin{figure}[!htb]
    \centering
    \includegraphics[width=0.9\linewidth]{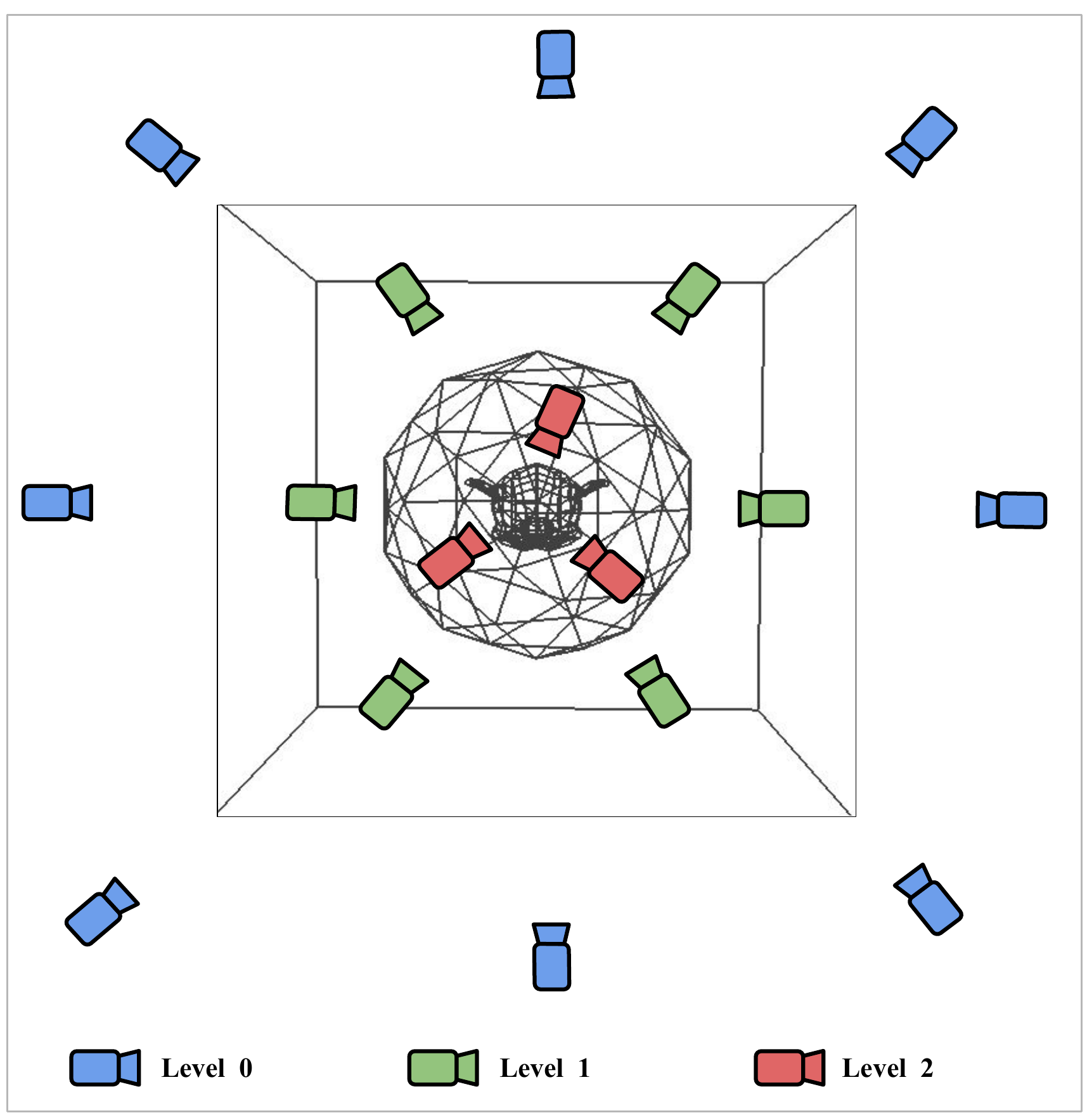}
    \caption{Top, wireframe view of a multi-layered stratified scene with three levels (monkey head inside sphere inside a cube).
    The camera colors indicate views of a specific level.
    \emph{Strata-NeRF} enables high-quality reconstruction of such stratified scenes using a single neural network.
    }
    \label{fig:teaser_image}
\end{figure}

Novel view synthesis is an ill-posed problem widely encountered in various areas such as augmented reality~\cite{kaneko2022ar, li2022rt}, virtual reality~\cite{deng2022fov}, etc.  A paradigm change for solving these kinds of problems was brought by the introduction of Neural Radiance Fields (NeRF)~\cite{nerf}. NeRFs are neural networks that take in the spatial coordinates and camera parameters as input and output the corresponding radiance field. Earlier version of NeRFs enable the generation of high-fidelity novel views for bounded scenes, significantly improving over existing techniques like Structure From Motion~\cite{schonberger2016structure}. Further, the capability of NeRFs have been recently extended to model unbounded scenes by Mip-NeRF 360~\cite{barron2022mip360}. This enabled NeRFs to model complex real-world scenes, where the scene content can exist at any distance from the camera.

However, similar to unboundedness in scenes, hierarchies in scenes are also natural. For example, images captured in a house can be categorized into images captured outside and inside across various rooms. Modelling such hierarchical scenes jointly for all levels through a NeRF could be particularly useful in cases of Virtual Reality applications. As it would not require switching to a different NeRF for each level,  reducing memory requirement and latency in switching. Further, as the different hierarchies of a scene usually share texture and architectural commonalities, it could lead to effective knowledge sharing and reduce the requirement of training independent models. For tackling the above novel objective, we introduce a paradigm of scenes that can be deconstructed into several tiers, termed \textit{``Stratified Scenes''}.  A ``stratified'' scene has several levels or groupings of structure (Figure~\ref{fig:teaser_image}). In our work, we first propose a synthetic dataset of stratified scenes, i.e. scenes having multiple levels. This dataset comprises scenes from two categories: (\RomanNumeral{1}) Simpler geometry, such as spheres, cubes, or tetrahedron meshes, and (\RomanNumeral{2}) Complex geometry, which closely emulates a real-world setup. 
\begin{figure}[!t]
    \centering
    \includegraphics[width=\linewidth]{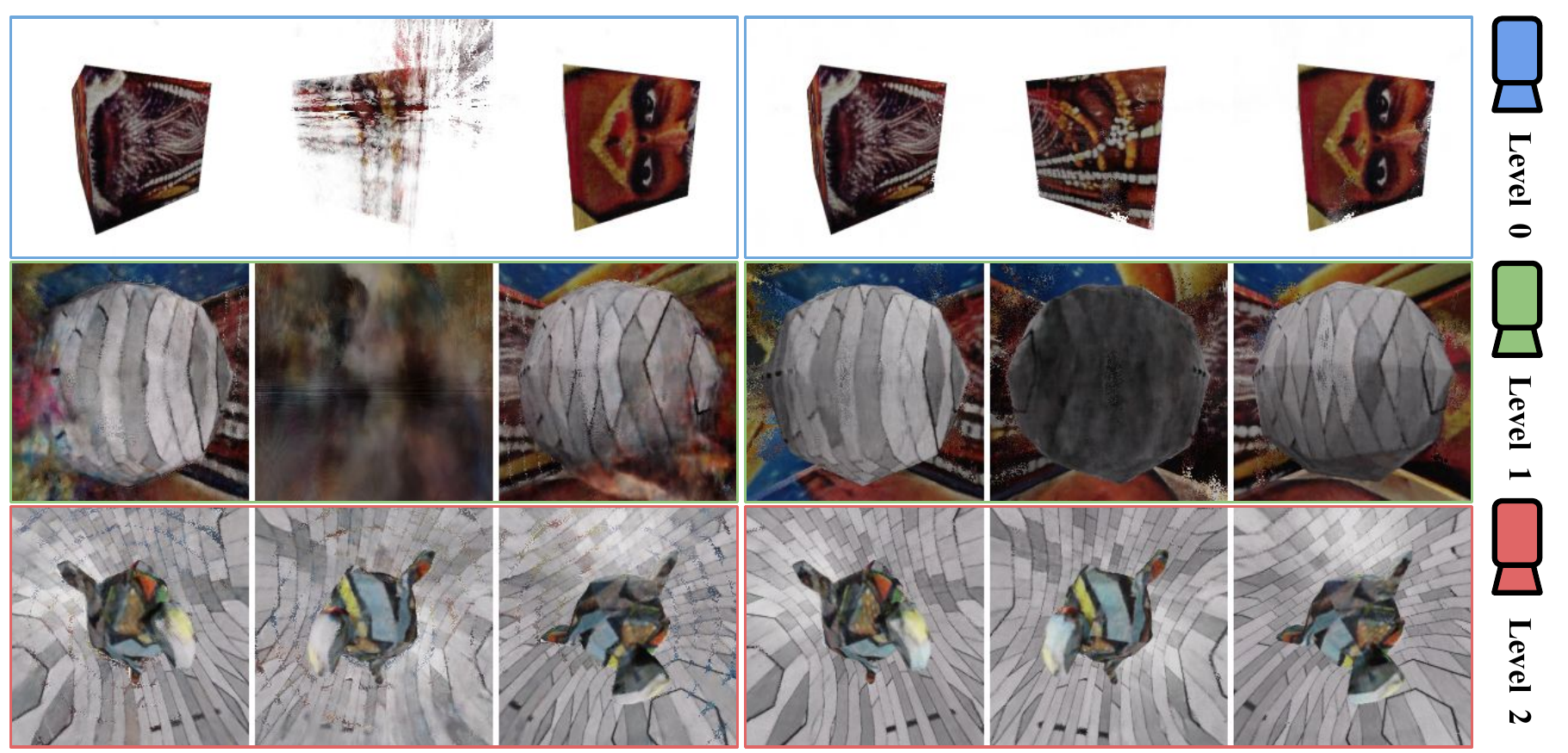}
    \caption{Novel views for stratified scene in Figure \ref{fig:teaser_image}, from {Mip-NeRF 360}~\cite{barron2022mip360}  \textit{(left)} and our method \textit{``Strata-NeRF''} \textit{(right)}.
    Existing methods struggle to capture stratified scenes with a single network while ours produces sharp results.
    }
    \label{fig:teaser_image_res}
\end{figure}

\begin{figure}[!t]
    \centering
    \includegraphics[width=\linewidth]{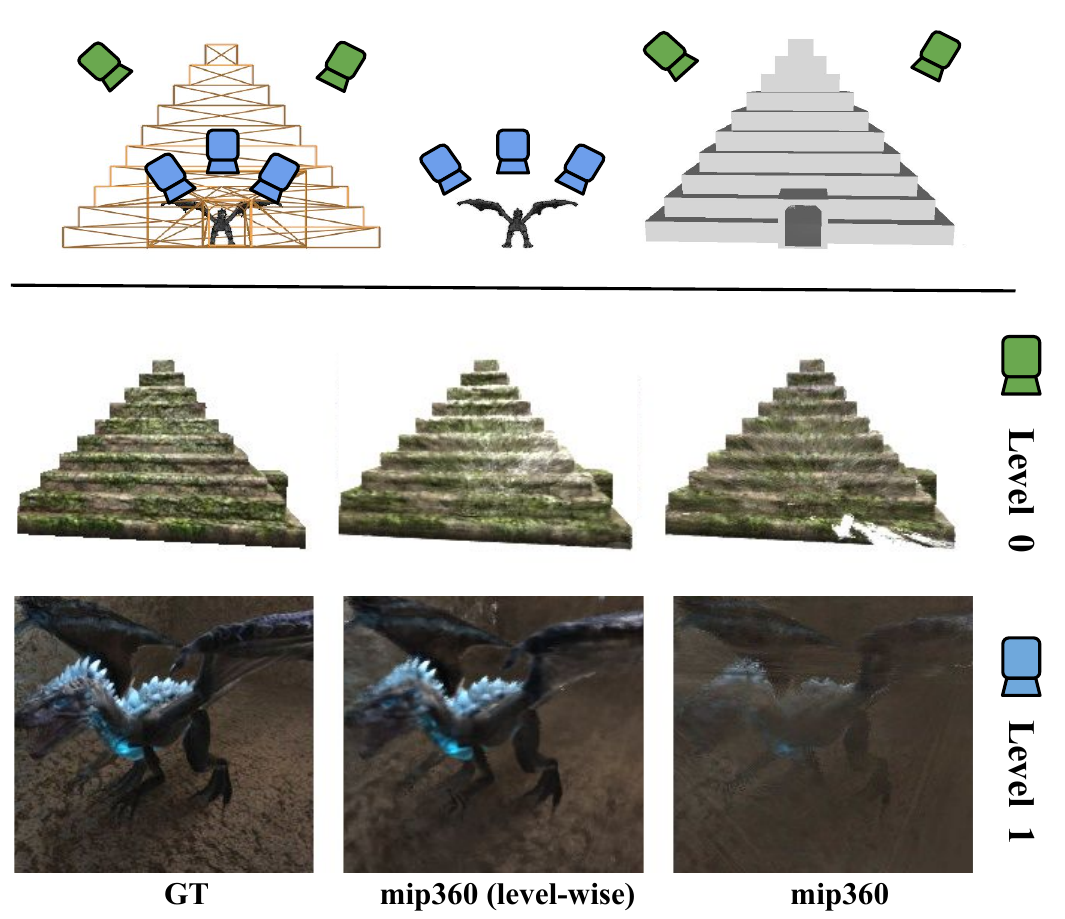}
    \caption{Analysis on ``Dragon in pyramid'' scene. The top row shows the layout of the levels in 3D scene. Observe that baseline works fine on the scenes when trained individually. Artefacts occur when the baseline is trained on views from the entire scene.}
    \label{fig:motivation_figure}
\end{figure}

On such datasets, we find methods such as Mip-NeRF 360 perform well for each level of the hierarchy independently, but produce unsatisfactory results when images from all hierarchical levels are used together for training (Figure~\ref{fig:motivation_figure}). This can be attributed to the continuous nature of NeRFs, which is unsuitable for modelling the sudden changes in scenes with shifts in hierarchical levels. Hence, in this work, we introduce \emph{Strata-NeRF} that explicitly aims to model the hierarchies by conditioning~\cite{jang2021codenerf, park2021nerfies, park2021hypernerf, yu2021pixelnerf, rebain2022lolnerf} the NeRF on Vector Quantized (VQ) latents. The VQ latents enable the modelling of discontinuities and sudden changes in the scene, as they are discrete and less correlated with others~\cite{van2017neural}. In practice, the VQ conditioning is achieved by introducing two lightweight modules: the ``Latent Generator'' module that compresses the implicit information in encoded 3D positions to generate VQ latent code, which is directed through the ``Latent Routing'' module to condition various layers of radiance field. The additional parameters introduced through these modules are significantly less than training an independent NeRF model for each level, leading to a significant reduction in memory.

For evaluating the proposed \emph{Strata-NeRF} we first test on the proposed synthetic \textit{Stratified Scenes} dataset, where we find that \textit{Strata-NeRF} learns the structure in scenes across all levels. In contrast, other baselines produce cloudy and sub-optimal novel views (Figure~\ref{fig:teaser_image_res}). Further, to test the generalizability of the proposed method on real-world scenes, we utilize the high-resolution RealEstate10K dataset. We find that \emph{Strata-NeRF} significantly outperforms other baselines  and produces high-fidelity novel views without artifacts compared to baselines. This is also observed quantitatively through improvement in metrics, where it establishes a new state-of-the-art. In summary,

\begin{itemize}
\setlength{\itemsep}{0pt}
    
    \item  We first introduce the task of implicit representation for 3D stratified (hierarchial) scenes using a single radiance field network. For this, we introduce a novel synthetic dataset comprising of scenes ranging from simple to complex geometries.
    \item For implicit modelling of the stratified scenes, we propose \emph{Strata-NeRF}, which conditions the radiance field based on discrete Vector-Quantized (VQ) latents to model the sudden changes in scenes due to change in hierarchical level (i.e. strata). 
    \item  \emph{Strata-NeRF} significantly outperforms the baselines across the synthetic dataset and generalizes well on the real-world scene dataset of RealState10k.  
\end{itemize}

\section{Related Work}
Generating photo-realistic novel views from densely sampled images is a classical problem. Earlier methods solved this issue using light-field-based interpolation techniques~\cite{davis2012unstructured, gortler1996lumigraph, levoy1996light}. These techniques interpreted the input images as 2D slices of a 4D function - the light field. The only caveat in these methods is their overreliance on dense views. Another popular technique is Structure From Motion (SFM) which reconstructs 3D structure of a scene or an object by using a sequence of 2D images. We suggest readers to read survey papers ~\cite{schonberger2016structure, ozyecsil2017survey} to understand SFM methods in detail. Shum \etal~\cite{shum2000review} also provides an excellent review on traditional image based rendering techniques. 

\noindent \textbf{Neural Volume Reconstruction.} 
NeRF~\cite{nerf} has shown remarkable results in encoding the 3D geometry of a scene implicitly using the multi-layer perceptron (MLP). Specifically, it trains an MLP, which takes 3D position and a viewing direction to predict colour and occupancy. Many papers have extended this idea to solve different scenarios such as dynamic scenes, low-light scenes, synthesis from fewer views, accelerating the performance etc. Mip-Nerf~\cite{barron2021mip} mitigates the problem of aliasing when a novel view is generated at a different resolution. MVSNeRF~\cite{chen2021mvsnerf} generalizes across all the scenes and optimizes the geometry and radiance field using only a few views. NerfingMVS~\cite{wei2021nerfingmvs} utilizes conventional SFM reconstruction and learning-based priors to predict the radiance field. UNISURF~\cite{oechsle2021unisurf} combines implicit surface models and radiance fields to render both surface and volume rendering.

AR-NeRF~\cite{kaneko2022ar} replaced pin-hole based camera ray-tracing with aperture camera based ray-tracing. DiVeR~\cite{wu2022diver} uses a voxel based representation to learn the radiance field, Mip-NeRF 360~\cite{barron2022mip360} improves view synthesis on the unbounded scenes and also proposed an online distillation scheme which significantly reduced the training and inference time. Neural Rays~\cite{liu2022neural} solves the occlusion problem by predicting the visibility of the 3D points in their representation. Scene Representation Transformers~\cite{sajjadi2022scene} uses Vision Transformers ~\cite{dosovitskiy2020image} to infer latent representations to render the novel views. Further, many methods~\cite{liu2020neural, garbin2021fastnerf, reiser2021kilonerf, yu2021plenoctrees, sun2022direct, hu2022efficientnerf, wang2022fourier} have been proposed to improve the slow training and inference time for neural radiance field based methods. Despite many works, no work has focused on modelling the \emph{stratified} scenes. 

\noindent \textbf{NeRF Extensions.}  Relighting discusses how to model different types of light and then using this model to re-light a scene~\cite{martin2021nerf, bi2020neural, srinivasan2021nerv, verbin2022ref, guo2022nerfren}. Breaking the myth that radiance field can only be used in small and bounded scenes, recent methods~\cite{tancik2022block, turki2022mega, rematas2022urban} have scaled it to large-scale city scenes. Another line of work focuses on  modelling the dynamic scenes with presence of moving objects~\cite{park2021nerfies, xian2021space, li2021neural, pumarola2021d, du2021neural, tretschk2021non, gao2021dynamic, park2021hypernerf} through NeRFs.   

\noindent \textbf{Neural Radiance Fields and Latents.} Recently, a lot of methods have made use of the latents to bring generative capabilities to neural radiance fields. GRAF~\cite{schwarz2020graf} uses disentangled shape and appearance latent codes to generalize on an object category. For viewpoint invariance, they used typical GAN based training. Pi-GAN~\cite{chan2021pi} uses volumetric rendering equations for consistent 3D views in a generative framework. Pixel-NeRF~\cite{yu2021pixelnerf} learns a scene prior to generalize across different scenes. GSN~\cite{devries2021unconstrained} decomposes the radiance field of a scene into local radiance fields by conditioning on a 2D grid of latent codes. Code-NeRF~\cite{jang2021codenerf} learns the variation of object shapes and textures across by learning separate latent embeddings. LOLNeRF~\cite{rebain2022lolnerf} uses a shared latent space which conditions a neural radinace field to model shape and appearance of a single class. PixNerF ~\cite{cai2022pix2nerf} extends Pi-GAN~\cite{chan2021pi} and  maps images to a latent manifold allowing object-centric novel views given a single image of an object. NeRF-W~\cite{Martin-Brualla_2021_CVPR} optimizes latent codes to model the scene variations to produce temporally consistent novel view renderings. In contrast to these methods, we propose conditioning NeRF on learnable Vector Quantized latents.

\textbf{Vector Quantized Variational Autoencoders (VQ-VAE)~\cite{van2017neural}}:  VQ-VAE uses vector quantization to represent a discrete latent ditribution. VQ-VAE has shown applications in Image Generation~\cite{razavi2019generating, peng2021generating}, speech and audio processing~\cite{gregor2018temporal, wang2019vector}. Further, it's extension like VQ-VAE2~\cite{razavi2019generating} uses hierarchical latent space for high-quality generation. 

\section{Preliminaries}
\label{sec:prelims}
NeRF represents a scene as an implicit function  $f : (X,d) \rightarrow (c,\sigma)$ which maps a 3D position $X=(x,y,z)$ and $d=(\theta, \phi)$ to a color $c=(r,g,b)$ and occupancy density $\sigma$. An MLP parametrizes this implicit function $f$. Before sending the inputs $X$ and $d$ through the network, a positional encoding is used to project them in a high dimensional space~\cite{tancik2020fourier}. Finally, the volume rendering~\cite{kajiya1984ray} procedure enables NeRF to represent scenes with photo-realistic rendering from novel camera viewpoints.

\noindent \textbf{Volume Rendering.} At the crux of NeRF lies the volume rendering equation. A ray $r(t) = o + td$ is cast from the camera center $o$ through the pixel along direction $d$. The pixel’s color value is estimated by integrating along the ray $r(t)$ as described in Eq.~\ref{eq:volume_rendering_eq} 
\begin{equation}
    c(r) = \int_{t_n}^{t_f} T(t)\sigma(r(t))c(r(t), d) \,dt \
    \label{eq:volume_rendering_eq}
\end{equation}
where transmittance $T(t) = exp(-\int_{t_n}^{t} \sigma(r(s)) \,ds) $ is the probability that a ray passes unhindered from the near plane ($t_n$)  to plane ($t$) and use this probability to integrate till far plane ($t_f$).
In Mip-NeRF~\cite{barron2021mip}, a ray $r(t)$ is divided into intervals $T_i = [t_i, t_{i+1})$ which corresponds to a conical frustum. For each interval $T_i$, it computes the mean and variance $(\mu, \Sigma)$ and uses it for integrated position encoding as illustrated in Eq. \ref{eq:mip_ipe}.
\begin{equation}
    \gamma(\mu,\Sigma) = \Bigg\{  \left[ sin(2^l \mu)exp(-2^{2l-1}diag(\Sigma)) \atop                cos(2^l\mu)exp(-2^{2l-1}diag(\Sigma))\right] \Bigg\}^{L-1}_{0}
    \label{eq:mip_ipe}
\end{equation}
This solves the aliasing issue in the original NeRF. Mip-NeRF 360~\cite{barron2022mip360} proposed coarse-to-fine online distillation for proposal sampling, which efficiently reduces the training time as the proposed MLP only predicts density. They also proposed ray parametrization and regularisation techniques to alleviate hanging artifacts in unbounded scenes. \textit{We'll refer Mip-NeRF 360~\cite{barron2022mip360} as mip360 in all our discussions.} We choose mip360~\cite{barron2022mip360} as the baseline for all our experiments. 

\section{Motivation}
\label{sec:motivation}
\begin{table}[!t]
\caption{A quantitative comparison of mip360 (level-wise) and mip360 (all views) on ``Dragon in pyramid'' scene.}
\label{tab:motivation_table}
\begin{adjustbox}{width=\linewidth}
\begin{tabular}{cccc|ccc}
\toprule
 & \multicolumn{3}{c|}{\textbf{Level 0}}          & \multicolumn{3}{c}{\textbf{Level 1}}           \\ \midrule
 & \textbf{PSNR $\uparrow$} & \textbf{SSIM $\uparrow$} & \textbf{LPIPS $\downarrow$} & \textbf{PSNR $\uparrow$} & \textbf{SSIM $\uparrow$} & \textbf{LPIPS $\downarrow$} \\ \midrule
\textbf{mip360 (level-wise)} & \textbf{31.5390} & \textbf{0.9181} & \textbf{0.1304} &  \textbf{29.8560} & \textbf{0.8133} & \textbf{0.3484} \\
\textbf{mip360}              & 30.8847 & 0.9006 & 0.1367  & 24.3876 & 0.7055 & 0.5163 \\
\bottomrule
\end{tabular}
\end{adjustbox}
\end{table}

The majority of real-world scenarios are stratified with multiple levels. For example, a commodity store has exterior and interior structures. This work addresses an essential question for such stratified scenes: \textit{Can a single radiance field learn such hierarchical scenes?} This section introduces and discusses our observations on one such stratified scene: ``Dragon in Pyramid'', as illustrated in Figure \ref{fig:motivation_figure}. The outer structure of ``Dragon in Pyramid'' is a Mayan pyramid that has a dragon inside it. To validate our claim, we first train the baseline model on each level, i.e., on outer pyramid views and inner views (focusing dragon) independently. We refer to these separately trained models as\textit{ mip360 (level-wise)}. Then, we train a single \textit{mip360}model using the outer and inner views for the scene. The term ``level" in our work refers to each level in a stratified scene. In the scene depicted in Figure \ref{fig:motivation_figure}, level $0$ denotes the pyramid's outer construction, while level $1$ denotes the pyramid's interior structure, which contains a dragon.

Table \ref{tab:motivation_table} shows that the baseline model performs remarkably well when trained separately on each level. In comparison, the metric values for the baseline model trained jointly on both levels of stratified scene declines. PSNR at level $1$ is $24.39 \, dB$, a $5.47 \,dB$ reduction compared to mip360 (level-wise). Similarly, performance in level $0$ has declined, but less dramatically than in the inner level. This pattern is observed across all metrics. Furthermore, the qualitative results illustrated in Figure \ref{fig:motivation_figure} backs up the quantitative study's findings. Figure \ref{fig:motivation_figure} indicates that mip360 (level-wise) generates novel views on par with the ground truth. However, shown in Figure \ref{fig:motivation_figure}, the jointly trained model has white artifacts on the pyramid's outer structure and haziness in front of the dragon inside the pyramid. This demonstrates that current radiance field networks have issues while learning a 3D representation of a stratified scene. We perform a similar experiment for a RealEstate10K scene in Appendix ~\refsupp{E.1} in the supplmentary material. 

\begin{figure*}[!t]
 \centering
    \includegraphics[width=\linewidth]{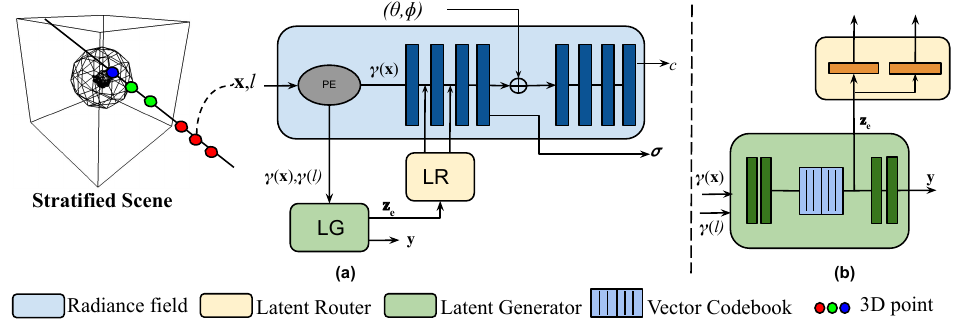}
    \caption{For each $3D$ point along the projected ray, we generate a latent code using our ``Latent generator" module. The generated latent code is routed to the MLP using ``Latent Router". Vector Codebooks learn the discrete distribution of positionally encoded $3D$ points. (a) Our model's end-to-end architecture; (b) components of the ``Latent Generator" and ``Latent Router" blocks.}
    \label{fig:architecture}
\end{figure*}

\section{Method}
\label{sec:method}
This section describes our method : \emph{Strata-NeRF} for  stratified scenes. We generate latent codes with the latent generator described in Section \ref{subsec:latent_generator}. This latent code is fed into the radiance field architecture through the latent router, described in Section \ref{subsec:latent_router}. Figure \ref{fig:architecture} depicts the overall architecture of Strata-NeRF. We adopt the base neural radiance field architecture proposed in mip360 ~\cite{barron2022mip360}.

\subsection{Latent Generator}
\label{subsec:latent_generator}

A latent space reflects the scene's ``compressed" representation. It has been shown in various works that this space has rich properties. VQ-VAE~\cite{van2017neural} learns a codebook to model the discrete distribution of the latent space of a variational-autoencoder. The encoder's output is compared to all of the vectors in the codebook. The nearest vector is fed into the decoder as input. Since most data in the world is discrete, VQ based models have been highly successful in image generation~\cite{esser2021taming}, speech encoding~\cite{van2017neural}, and other applications. In a stratified scene, the definition of level is also discrete.Hence, our method employs VQ-VAE as a latent generator because of their proven success in representing discrete distributions.

We use Integrated Positional Encoded (IPE)~\cite{barron2022mip360} $\gamma(\mathbf{x})$  as input to our latent generator. We encode $\gamma(\mathbf{x})$ and then search the codebook for the closest vector. After that, the closest vector from the codebook is used to condition the radiance field network. Specifically, $\gamma(\mathbf{x})$ is passed through a set of two hidden layers to generate an encoded input $\mathbf{z}$. The encoded latent code $\mathbf{z}$ is then passed through the quantizer bottleneck to determine the quantized latent code $\mathbf{z_e}$, where $\mathbf{z_e} \in E$; where $E \in {R}^{N \times D}$ is the codebook; $N$ is the number of vectors in the codebook, and $D$ is the dimension of the latent space. $\mathbf{z_e}$  is then supplied into the decoder network, which consists of two hidden layers, to yield $\mathbf{y}$ as the reconstructed output of $\gamma(\mathbf{\mathbf{x}})$. The quantized latent $\mathbf{z_e}$  is also sent into the radiance field network through the ``Latent Router'' block. Loss for this variational autoencoder (VAE) block is defined as follows:
\begin{equation}
\label{eq:vae_loss}
    \mathcal{L}_{vq} = ||\gamma(\mathbf{x}) - \mathbf{y}||_2^2 + ||sg(\mathbf{z_e}) - \mathbf{z}||_2^2 + \beta ||\mathbf{z_e} - sg(\mathbf{z})||_2^2 
\end{equation}
The ``Latent Generator''  module based on VAE is jointly trained with the NeRF through backpropagation. 

\subsection{Latent Router}
\label{subsec:latent_router}
The Latent Router block is inspired by the CodeNeRF architecture~\cite{jang2021codenerf}, in which shape and texture latent codes are sent to the NeRF MLP through a residual connection. In our architecture, the quantized latent codes $z_e$ that are generated in the ``Latent Generator'' block are input to the Radiance field after passing through an MLP layer in the Latent Router as shown in Figure \ref{fig:architecture}. 

\subsection{ Training Strata-NeRF} 
For training Strata-NeRF, we utilize the losses suggested by mip360~\cite{barron2022mip360} as we use a similar radiance field design. $\mathcal{L}_{recon}(c(r,t),c^{*}(r))$ denotes the reconstruction loss between the estimated colour along a ray and the actual colour value. $\mathcal{L}_{dist}(s,w)$ is the distortion loss where $s$ is the normalized ray distances and $w$ is the weight vector. Note that we don't alter anything in the proposal MLP. More details are provided in mip360~\cite{barron2022mip360}.  The total loss for Strata-NeRF is given as:
\begin{equation}
\label{eq:total_loss}
    \mathcal{L}_{total} = \mathcal{L}_{recon}(c(r,t),c^{*}(r)) + \lambda_{1} \mathcal{L}_{dist}(s,w) +  \lambda_{2} \mathcal{L}_{vq}
\end{equation}
We use $\lambda_{1}=0.01$, $\lambda_{2}=0.1$ and $\beta = 1.0$ across all our experiments, as they work robustly~\cite{barron2022mip360} for \emph{Strata-NeRF}.

\begin{table}[!t]
\caption{Characteristic Comparison of the proposed methods}
\label{tab:characteristic_table}
\begin{adjustbox}{width=\linewidth}
\begin{tabular}{llll}
\hline
\textbf{Method} &
  \textbf{\begin{tabular}[c]{@{}l@{}}Discrete \\ Representation\end{tabular}} &
  \textbf{\begin{tabular}[c]{@{}l@{}}Photometric \\ Losses\end{tabular}} &
  \textbf{\begin{tabular}[c]{@{}l@{}}VAE \\ loss\end{tabular}} \\ \hline
 \textbf{NeRF~\cite{nerf}}           & \xmark & \cmark  & \xmark \\
\textbf{mip360~\cite{barron2022mip360}}           & \xmark & \cmark  & \xmark \\
\textbf{Plenoxel\cite{yu2021plenoxels}}   & \cmark & \cmark  & \xmark \\
\textbf{Instant-NGP\cite{muller2022instant}}   & \cmark & \cmark  & \xmark \\
\textbf{TensoRF\cite{chen2022tensorf}}   & \cmark & \cmark  & \xmark \\
\textbf{Ours}       & \cmark & \cmark  & \cmark \\
\hline
\end{tabular}
\end{adjustbox}
\end{table}
\section{Experiments}
We discuss implementation details in Section ~\ref{subsec:implementation_details}. Section ~\ref{subsec:dataset} discusses the dataset used for evaluating our method with other baselines. In Section ~\ref{subsec:evaluation}, we present quantiative and qualitative comparison with the baseline methods. Additionally, we discuss the ablations for the proposed method. 

\label{sec:experiments}
\subsection{Implementation Details}
\label{subsec:implementation_details}
Our method builds on mip360~\cite{barron2022mip360} as the base radiance field. We use a latent generator network which consists of an encoder-decoder architecture and a vector-codebook. The encoder has two linear layers of hidden size  $48$, and the decoder has one linear layer of hidden size $96$. The output dimension of our decoder matches the output from Integrated Positional Encoding (IPE) block. The size of our codebook is $1024$, and the dimension of each vector in the codebook is $48$. We condition the neural radiance field through the latent generated after the quantization step in the latent generator. We use a Latent routing module consisting of two linear layers of hidden-size $256$. As illustrated in Figure \ref{fig:architecture}, the output of the linear layer in the routing module conditions the first two layers of the radiance field network.   We employ the losses outlined in Section \ref{sec:method}. On each scene, we train our approach for $150k$ iterations. We use Adam~\cite{kingma2014adam} optimizer with a learning rate of $1e^{-6}$. Further details are provided in supplementary material.
\begin{figure}[!t]
    \centering
    \includegraphics[width=\linewidth]{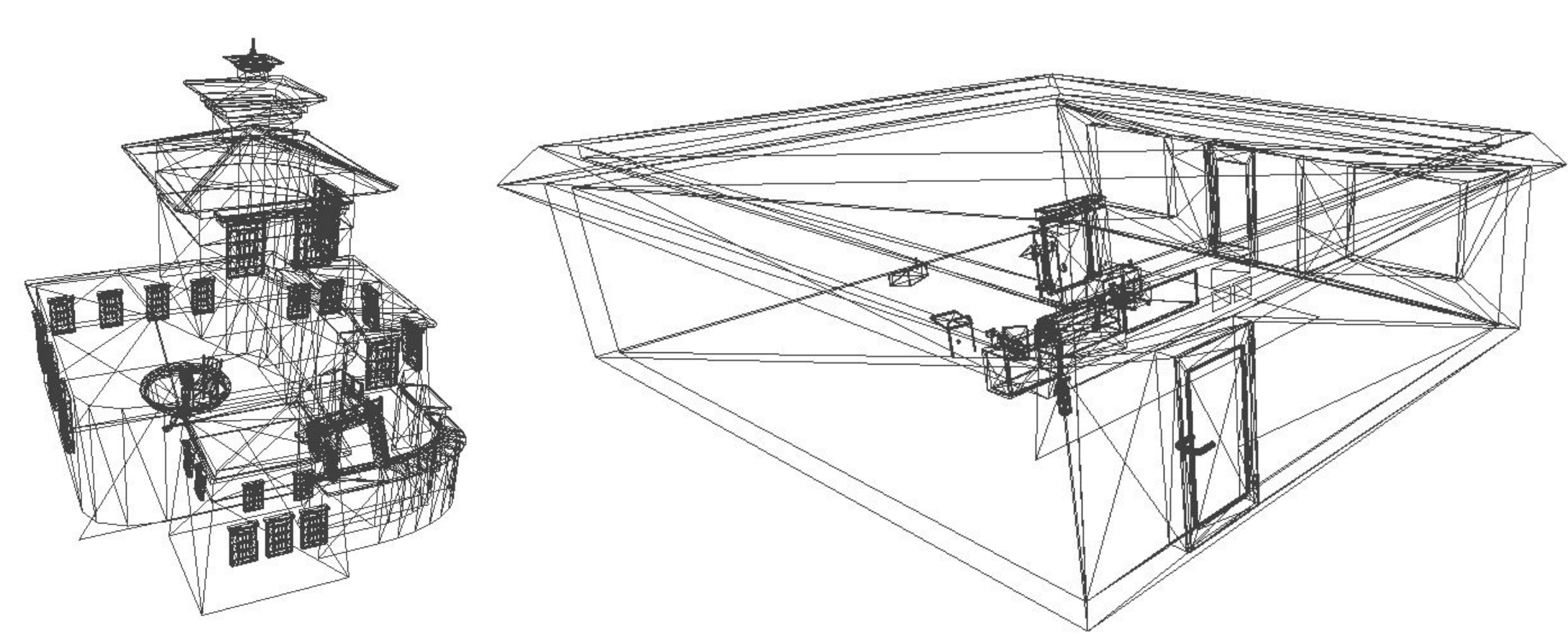}
    \caption{Skeleton mesh of the stratified scenes : Bhutanese House and Coffee Shop. More details are in the supplementary material. }
    \label{fig:dataset_overview}
\end{figure}
\subsection{Evaluation Dataset}
\label{subsec:dataset}

\begin{table*}[!t]
\caption{Quantitative evaluation on test-set against baselines discussed in Section \ref{subsec:implementation_details}. Each column is depicts the \textbf{best} and \underline{second best}.}
\label{tab:characteristic_table_1}
\centering
\begin{adjustbox}{width=\linewidth}
\begin{tabular}{c|ccc|ccc|ccc|ccc} 
\hline
                            & \multicolumn{12}{c}{\textbf{Cube-Sphere-Monkey}~}                                           \\ 
\cline{2-13}
                            & \multicolumn{3}{c}{\textbf{Level 0}} & \multicolumn{3}{c}{\textbf{Level 1}} & \multicolumn{3}{c|}{\textbf{Level2}} & \multicolumn{3}{c}{\textbf{Total}}  \\ 
\cline{2-13}
                            & \textbf{PSNR $\uparrow$} & \textbf{SSIM $\uparrow$} & \textbf{LPIPS $\downarrow$}         & \textbf{PSNR $\uparrow$} & \textbf{SSIM $\uparrow$} & \textbf{LPIPS $\downarrow$}         & \textbf{PSNR $\uparrow$} & \textbf{SSIM $\uparrow$} & \textbf{LPIPS $\downarrow$}         & \textbf{PSNR $\uparrow$} & \textbf{SSIM $\uparrow$} & \textbf{LPIPS $\downarrow$}               \\ 
\hline
\textbf{Nerf}  ~\cite{nerf}                     & \underline{28.3314} & \underline{0.9383} & \underline{0.1034}   &                    18.1806 &                    0.4976 &                    0.4981 &                    22.1178 &                    0.5995 &                    0.3825 &                    22.8766 &                    0.6784 &                     0.3280                              \\
\textbf{mip360}  ~\cite{barron2022mip360}           & 28.3149                    & 0.9298                   & 0.1156 & \underline{19.0443} &                    \underline{0.5343} &                    \underline{0.4930} & \underline{24.9136} & \underline{0.7326} & \underline{0.3245} & \underline{24.0909} & \underline{0.7322} &                     \underline{0.3110}                         \\
\textbf{Plenoxels~\cite{yu2021plenoxels}}        & 25.3547   & 0.9169   & 0.1238                   &                    13.1148 & 0.3320 & 0.6895 &                    21.5568 &                    0.6523 &                    0.3803 &                    20.0087 &                    0.6337 & 0.3979                        \\ 
\textbf{Instant-NGP~\cite{muller2022instant}}        & 28.2104   & 0.9168   & 0.1123                   &                    14.3648 & 0.1830 & 0.7216 &                    17.6914 &                    0.2744 &                    0.5997 &                    20.0889 &                    0.4581 & 0.4779                           \\
\textbf{TensoRF~\cite{chen2022tensorf}}        & \textbf{32.0077}   & \textbf{0.9532}   & \textbf{0.0692}                   &                    13.7487 & 0.1537 & 0.7106 &                    13.0075 &                    0.2496 &                    0.6886 &                    19.5880 &                    0.4521 & 0.4894                          \\
\textbf{Ours}           & 26.9335                    & 0.9298                    & 0.1255                    & \textbf{25.7088}   & \textbf{0.7738}   & \textbf{0.2959}   & \textbf{26.1912}   & \textbf{0.8172}   & \textbf{0.2549}   & \textbf{26.2778}   & \textbf{0.8403}   & \textbf{0.2254}                          \\
\hline
                        &      &      &               &      &      &               &      &      &               &      &      &       \\
\hline
                            & \multicolumn{12}{c}{\textbf{Bhutanese House}~}                            \\ 
\cline{2-13}
                            & \multicolumn{3}{c}{\textbf{Level 0}} & \multicolumn{3}{c}{\textbf{Level 1}} & \multicolumn{3}{c|}{\textbf{Level 2}} & \multicolumn{3}{c}{\textbf{Total}}  \\ 
\cline{2-13}
                            & \textbf{PSNR $\uparrow$} & \textbf{SSIM $\uparrow$} & \textbf{LPIPS $\downarrow$}         & \textbf{PSNR $\uparrow$} & \textbf{SSIM $\uparrow$} & \textbf{LPIPS $\downarrow$}         & \textbf{PSNR $\uparrow$} & \textbf{SSIM $\uparrow$} & \textbf{LPIPS $\downarrow$}         & \textbf{PSNR $\uparrow$} & \textbf{SSIM $\uparrow$} & \textbf{LPIPS $\downarrow$}              \\ 
\hline
\textbf{Nerf} ~\cite{nerf} &                    11.4478 &                    0.6917 &                    0.3711 &                    17.1209 &                    0.5886 &                    0.7078 &                    18.3918 &                    0.6952 &                    0.6591 &                    15.6535 &                    0.6585 &                    0.5793             \\
\textbf{mip360 } ~\cite{barron2022mip360}                   &                    \underline{26.6240} & 0.9002 & 0.2062 & 24.5946 &                    \underline{0.7296} &                    0.4739 & \underline{29.4225} & \textbf{0.8577} & \underline{0.4156} & \underline{26.8804} &                    \underline{0.8291} &                    0.3652          \\
\textbf{Plenoxels~\cite{yu2021plenoxels} }       & 15.2205 &                    0.7752 &                    0.3052 &                    13.0386 & 0.4670   & 0.6703   &                    19.3050 & 0.5819   & 0.5886   &                    15.8547 & 0.6080   & 0.5214                    \\ 
\textbf{Instant-NGP~\cite{muller2022instant} }       & 23.9791 &                    \textbf{0.9217} &                    \textbf{0.1500} &                    \underline{24.7316} & 0.7009   & \underline{0.4237}   &                    27.6617 & 0.8136   & \textbf{0.3786}   &                    25.4575 & 0.8121   & \textbf{0.3174}                    \\
\textbf{TensoRF~\cite{chen2022tensorf} }       & 13.8880 &                    0.7607 &                    0.3142 &                    17.0244 & 0.4856   & 0.6421   &                    16.8170 & 0.6306   & 0.6332   &                    15.9098 & 0.6256   & 0.5298                   \\
\textbf{Ours }           & \textbf{27.6842}   & \underline{0.9046}   & \underline{0.2045}   & \textbf{24.9180}   & \textbf{0.7371} & \textbf{0.4616} & \textbf{29.4646}   & \underline{0.8575}                    & 0.4172                    & \textbf{27.3556}   & \textbf{0.8331} & \underline{0.3611}                 \\
\hline
\end{tabular}
\end{adjustbox}
\end{table*}

\begin{table*}[!t]
\caption{Quantitative evaluation on test-set against baselines discussed in Section \ref{subsec:implementation_details}. Each column is depicts the \textbf{best} and \underline{second best}.}
\label{tab:characteristic_table_1_part_2}
\centering
\begin{adjustbox}{width=\linewidth}
\begin{tabular}{c|ccc|ccc|ccc|ccc} 
\hline
                            & \multicolumn{12}{c}{\textbf{Coffee Shop}~}                                           \\ 
\cline{2-13}
                            & \multicolumn{3}{c}{\textbf{Level 0}} & \multicolumn{3}{c}{\textbf{Level 1}} & \multicolumn{3}{c|}{\textbf{Level2}} & \multicolumn{3}{c}{\textbf{Total}}  \\ 
\cline{2-13}
                            & \textbf{PSNR $\uparrow$} & \textbf{SSIM $\uparrow$} & \textbf{LPIPS $\downarrow$}         & \textbf{PSNR $\uparrow$} & \textbf{SSIM $\uparrow$} & \textbf{LPIPS $\downarrow$}         & \textbf{PSNR $\uparrow$} & \textbf{SSIM $\uparrow$} & \textbf{LPIPS $\downarrow$}         & \textbf{PSNR $\uparrow$} & \textbf{SSIM $\uparrow$} & \textbf{LPIPS $\downarrow$}               \\ 
\hline
\textbf{Nerf}  ~\cite{nerf}                     & 06.7446 &                    0.6197 &                    0.4698 &                    16.1398 &                    0.4915 &                    0.7982 &                    12.8889 &                    0.4213 &                    0.8158 &                    11.9244 &                    0.5108 &                    0.6946               \\
\textbf{mip360}  ~\cite{barron2022mip360}           &  26.2073 & 0.8825   & \textbf{0.1867} &                    27.0500 &                    0.8086 &                    0.3785 & \textbf{34.2023}   & \textbf{0.9362}   & \textbf{0.1950}   & 29.1532 &                    \underline{0.8757} & \underline{0.2534}              \\
\textbf{Plenoxels~\cite{yu2021plenoxels}}         &            19.3204         &                    0.7968 & 0.2579   & 12.3871   & 0.4044   & 0.6904   & 22.4325 & 0.6856 & 0.4585 &                    18.0467 & 0.6289   & 0.4689              \\ 
\textbf{Instant-NGP~\cite{muller2022instant}}        &            \underline{29.9425}         &                    \underline{0.9324} & \underline{0.0992}  & \underline{28.1040}   & \underline{0.8193}   & \underline{0.3452}   & 29.6574 & 0.8680 & 0.2621 &                    \underline{29.2347} & 0.8732   & \textbf{0.2355}             \\
\textbf{TensoRF~\cite{chen2022tensorf}}        &           \textbf{33.0337}         &                    \textbf{0.9435} & \textbf{0.0692}   & 19.3115   & 0.5331   & 0.6580   & 21.1852 & 0.7169 & 0.4594 &                    24.5102 & 0.7312   & 0.3955              \\
\textbf{Ours}           & 26.4499   & 0.8802 & \underline{0.1939}                    & \textbf{28.6392} & \textbf{0.8403} & \textbf{0.3450} & \underline{33.2692}                    & \underline{0.9254}                    & \underline{0.2243}                    & \textbf{29.4528}   & \textbf{0.8819} & 0.2544             \\
\hline
                        &      &      &               &      &      &               &      &      &               &      &      &                                \\
\hline
                            & \multicolumn{12}{c}{\textbf{Dragon In Pyramid}~}                            \\ 
\cline{2-13}
                            & \multicolumn{3}{c}{\textbf{Level 0}} & \multicolumn{3}{c}{\textbf{Level 1}} & \multicolumn{3}{c|}{\textbf{Level 2}} & \multicolumn{3}{c}{\textbf{Total}}  \\ 
\cline{2-13}
                            & \textbf{PSNR $\uparrow$} & \textbf{SSIM $\uparrow$} & \textbf{LPIPS $\downarrow$}         & \textbf{PSNR $\uparrow$} & \textbf{SSIM $\uparrow$} & \textbf{LPIPS $\downarrow$}         & \textbf{PSNR $\uparrow$} & \textbf{SSIM $\uparrow$} & \textbf{LPIPS $\downarrow$}         & \textbf{PSNR $\uparrow$} & \textbf{SSIM $\uparrow$} & \textbf{LPIPS $\downarrow$}              \\ 
\hline
\textbf{Nerf} ~\cite{nerf} &                                        14.6405 &                    0.6595 &                      0.3800 &                    20.8368 &                    0.6052 &                    0.6856         &   -   &   -   &  -             &                    17.7386 &                    0.6323 &                    0.5328       \\
\textbf{mip360 } ~\cite{barron2022mip360}                   &                     \underline{30.8758} & 0.9006   & 0.1367   & 24.3890 &                    \underline{0.7054} &                    0.5163        &   -   &   -   &   -            & \underline{27.6324} &                    \underline{0.8030} &                    0.3265        \\
\textbf{Plenoxels~\cite{yu2021plenoxels} }       &  13.0667   & 0.6247 & 0.4217 &                    14.5126 & 0.3572 & 0.6498               &   -   &   -   &      -         &                    13.7896 & 0.4910   & 05358              \\ 
\textbf{Instant-NGP~\cite{muller2022instant} }       &  23.9054   & \underline{0.9010} & \underline{0.0949} &                    \underline{24.7389} & 0.6594 & \underline{0.4664}               &   -   &   -   &      -         &                    24.3222 & 0.7802   & \underline{0.2807}            \\
\textbf{TensoRF~\cite{chen2022tensorf} }       &  \textbf{35.3015}   & \textbf{0.9632} & \textbf{0.0414} &                    19.5573 & 0.5221 & 0.6809               &   -   &   -   &      -         &                    27.4294 & 0.7427   & 0.3611              \\
\textbf{Ours }           &  29.4773                    & 0.8700                    & \underline{0.1699}                    & \textbf{26.1722}   & \textbf{0.7489}   & \textbf{0.4573}          &      -&  -    &          -     & \textbf{27.8248}   & \textbf{0.8095} & \textbf{0.3136}        \\
\hline
\end{tabular}
\end{adjustbox}
\end{table*}

Most of the radiance field methods evaluate their results on the synthetic (Blender) and real-world (LLFF) datasets proposed in NeRF~\cite{nerf}. These scenes either include a solitary object on a white background or a frontal view of a natural scene. According to our description of stratified scenes, these datasets has only one level. Even large-scale reconstruction datasets like TanksandTemples~\cite{Knapitsch2017} are not representative of our setting as they only have views either inside or outside of the structure. Similarly, Scannet~\cite{dai2017scannet} a dataset for real-world interior scenes, lacks the characteristics of a stratified dataset. Because of the direct unavailability of stratified scenes, we built our own dataset that replicates the intended ``stratified" scenario. We create a synthetic scene dataset using a mesh-editing software Blender~\cite{blender} and real scene dataset by altering RealEstate10K dataset which was proposed for the camera localization task.

The proposed synthetic dataset has two important variations based on: (a) the number of stratified levels and (b) the geometric complexity. We classify based on the geometry's complexity as follows: (a) \textit{Simple Scenes}: Stratified scenes using geometric components such as the sphere, cube, and so on; and (b) \textit{Complex Scenes}: Stratified scenes that mimic real-world scenes. For Simple Scenes, we leverage models and textures provided by Blender ~\cite{blender}. We utilized publicly available graphical models and composited them to create a real-world configuration for Complex scenes. For example, to design the \textit{``Coffee shop''} scene, we selected a building structure for the outer level and walls and glasses for the intermediate level structure. For the core level, we composited elements such as a cash register, coffee cups, and so on to simulate a real-world coffee-shop scene. To avoid photo-metric changes, we use fixed illumination. For each stratified level, the camera settings : field of vision and focal length are fixed. Each scene is rendered at $200 \times 200$ resolution. The camera viewpoint are sampled evenly from the curved surface of a hemisphere and then randomly divided into train, validation, and test sets. Inner objects in \textit{Simple Scenes} are rendered from the surface of a sphere. Figure \ref{fig:dataset_overview} depicts the proposed dataset's skeletal meshes. Further information on dataset is present in Appendix ~\refsupp{B} in the supplementary material.

\noindent \textbf{RealEstate10K dataset.} We extracted four scenes \textit{``Spanish Colonial Retreat in Scottsdale Arizona'', ``139 Barton Avenue Toronto Ontario'' , ``31 Brian Dr Rochester NY'' and ``7 Rutledge Ave Highland Mills''} from RealEstate10K dataset. We manually inspected and removed regions which had dynamic components in them. More details about converting RealEstaet10k dataset for our stratified setting is provided in Appendix ~\refsupp{C} the supplementary material.

\subsection{Evaluation}
\label{subsec:evaluation}
\begin{figure}[!t]
    \centering
    \includegraphics[width=\linewidth]{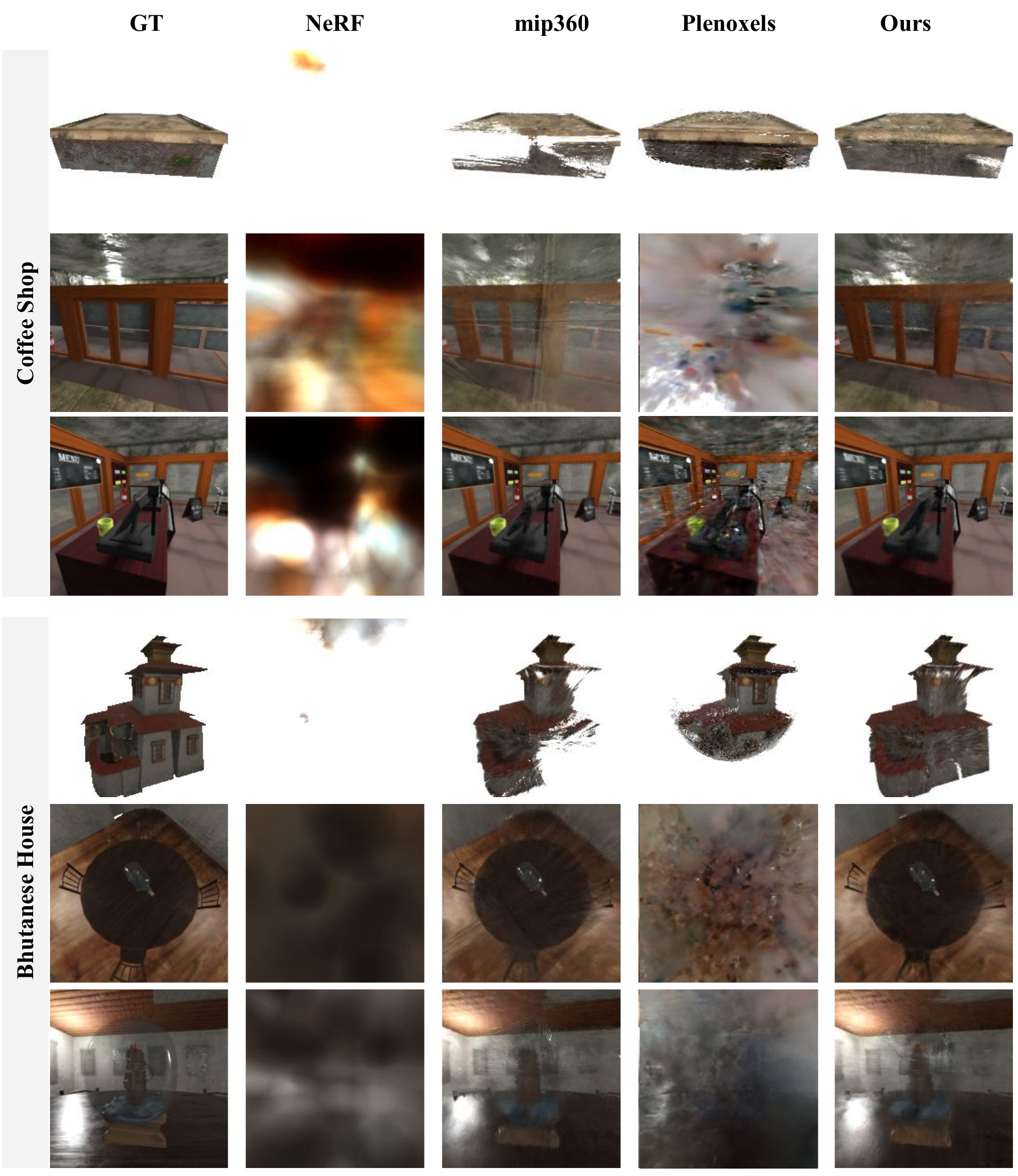}
    \caption{(From top to bottom) Qualitative results on the proposed synthetic datasets (Figure~\ref{fig:dataset_overview}). Each row represents a novel view from a level of the stratified scene. The ground-truth (GT) is shown in Column 1. Compared to baselines (Column 2-4), our method's (Column 5) renderings are more consistent to GT.}
    \label{fig:qual_results_1}
\end{figure}

\label{subsec:evaluation}
We present quantitative and qualitative analysis of \emph{Strata-NeRF} on the datasets described in Section~\ref{subsec:dataset}. 

\noindent \textbf{Baselines.} We compare our model with NeRF~\cite{nerf} , mip360~\cite{barron2022mip360}, Instant-NGP~\cite{muller2022instant}, TensoRF~\cite{chen2022tensorf} and Plenoxels~\cite{yu2021plenoxels}. We chose Plenoxels~\cite{yu2021plenoxels} for comparison because it uses sparse-voxel representation which already discretizes the continuous 3D space, which can be useful in stratified scenes. It is worth noting that the sizes of the synthetic scenes in our dataset differ. As a consequence, the authors' recommended configuration file did not produce the optimal results. As a result, we modified the configuration files for unbounded scenes released by the creators of mip360~\cite{barron2022mip360} to improve performance. For Instant-NGP~\cite{muller2022instant}, TensoRF~\cite{chen2022tensorf} and Plenoxels~\cite{yu2021plenoxels}, we change the hyperparameters like bound and scale as suggested in the official implementatinos. More information is in Appendix ~\refsupp{D} in the supplementary material. Table~\ref{tab:characteristic_table} provides an overview of baselines.

\noindent \textbf{Quantitative Results.} Table \ref{tab:characteristic_table_1} \& \ref{tab:characteristic_table_1_part_2} shows the average PSNR, SSIM~\cite{wang2004image} and LPIPS~\cite{zhang2018unreasonable} for each stratified level in unseen test views. We find that our method surpasses other methods across all metrics most of the time. The baseline mip360~\cite{barron2022mip360} works fine for the exterior structure but fails for the inner layers in the ``Cube-Sphere-Monkey" scene. \emph{Strata-NeRF}, on the other hand, offers superior metrics at all stratified levels. The baseline models do well in the outer scene but perform sub-optimally in the inner levels, especially in level 1. These outcomes demonstrate that our method outperforms the baseline models significantly.

Table \ref{tab:summary_real10} shows the summary of average PSNR and SSIM for all the levels in a scene for RealEstate10K dataset. In this case, we only compare our method with mip360 as it is the best performing one among others on the synthetic dataset.  We observe that our method outperforms the baseline method in all scenarios. Further, we present level-wise result for a specific scene in Table \ref{tab:six_layer_real10k}. We observe that for real datasets \textit{with increase in number of levels, the magnitude of performance improvement increases}, which demonstrates the effectiveness of the proposed approach. Further, we also compare Instant-NGP~\cite{muller2022instant} and TensoRF~\cite{chen2022tensorf} on a RealEstate10K scene in Appendix ~\refsupp{E.2} in the supplementary material. 

\begin{figure*}[!t]
    \centering
    \includegraphics[width=\linewidth]{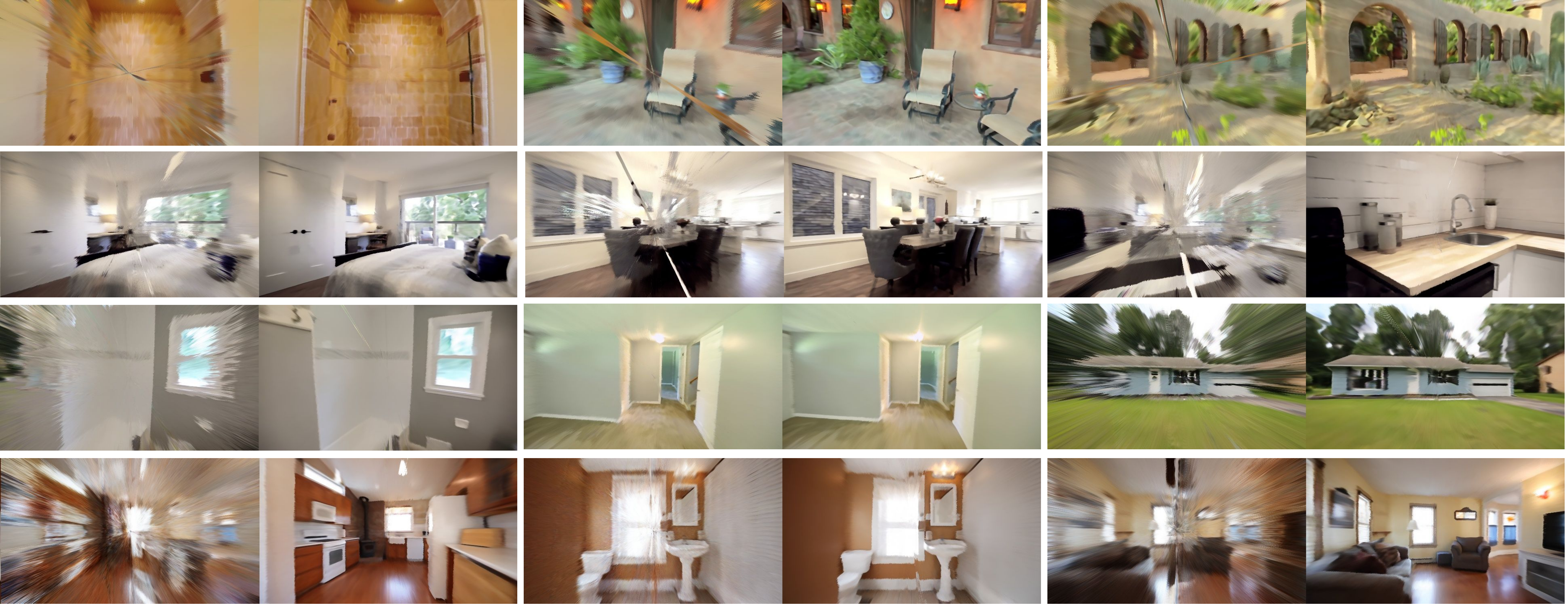}
    \caption{Qualitative comparison on Scenes from RealEstate10K dataset between mip360 (left image) and our method \emph{Strata-NeRF} (right image) in a pair. Each row represents a scene in RealEstate10K and each pair represents a level in that scene. Our method outperforms and produce good quality novel views compared to mip360.  }
    \label{fig:qual_real10k}
\end{figure*}
\begin{table}[!t]
\caption{Quantitative comparison of our model and baseline on \textbf{\textit{``139 Barton Avenue''}} scene of RealEstate10K dataset.}
\label{tab:six_layer_real10k}
\begin{adjustbox}{width=\linewidth}
\begin{tabular}{@{}cccccccc@{}}
\toprule
 &
  \textbf{Metrics} &
  \textbf{Level 0} &
  \textbf{Level 1} &
  \textbf{Level 2} &
  \textbf{Level 3} &
  \textbf{Level 4} &
  \textbf{Level 5} \\ \midrule
\multicolumn{1}{c|}{\multirow{2}{*}{\textbf{mip360~\cite{barron2022mip360}}}} &
  \multicolumn{1}{c|}{\textbf{PSNR $\uparrow$}} &
  18.086 &
  16.496 &
  24.459 &
  20.862 &
  17.479 &
  10.999 \\
\multicolumn{1}{c|}{} &
  \multicolumn{1}{c|}{\textbf{SSIM $\uparrow$}} &
  0.618 &
  0.595 &
  0.771 &
  0.702 &
  0.584 &
  0.409 \\ \midrule
\multicolumn{1}{c|}{\multirow{2}{*}{\textbf{Ours}}} & \multicolumn{1}{c|}{\textbf{PSNR $\uparrow$}} & \textbf{23.164} & \textbf{21.665} & \textbf{25.236} & \textbf{24.156} & \textbf{22.879} & \textbf{25.409} \\
\multicolumn{1}{c|}{} &
  \multicolumn{1}{c|}{\textbf{SSIM $\uparrow$}} &
  \textbf{0.826} &
  \textbf{0.757} &
  \textbf{0.789} &
  \textbf{0.791} &
  \textbf{0.753} &
  \textbf{0.782} \\ \bottomrule
\end{tabular}
\end{adjustbox}
\end{table}

\begin{figure}[!t]
    \centering
    \includegraphics[width=\linewidth]{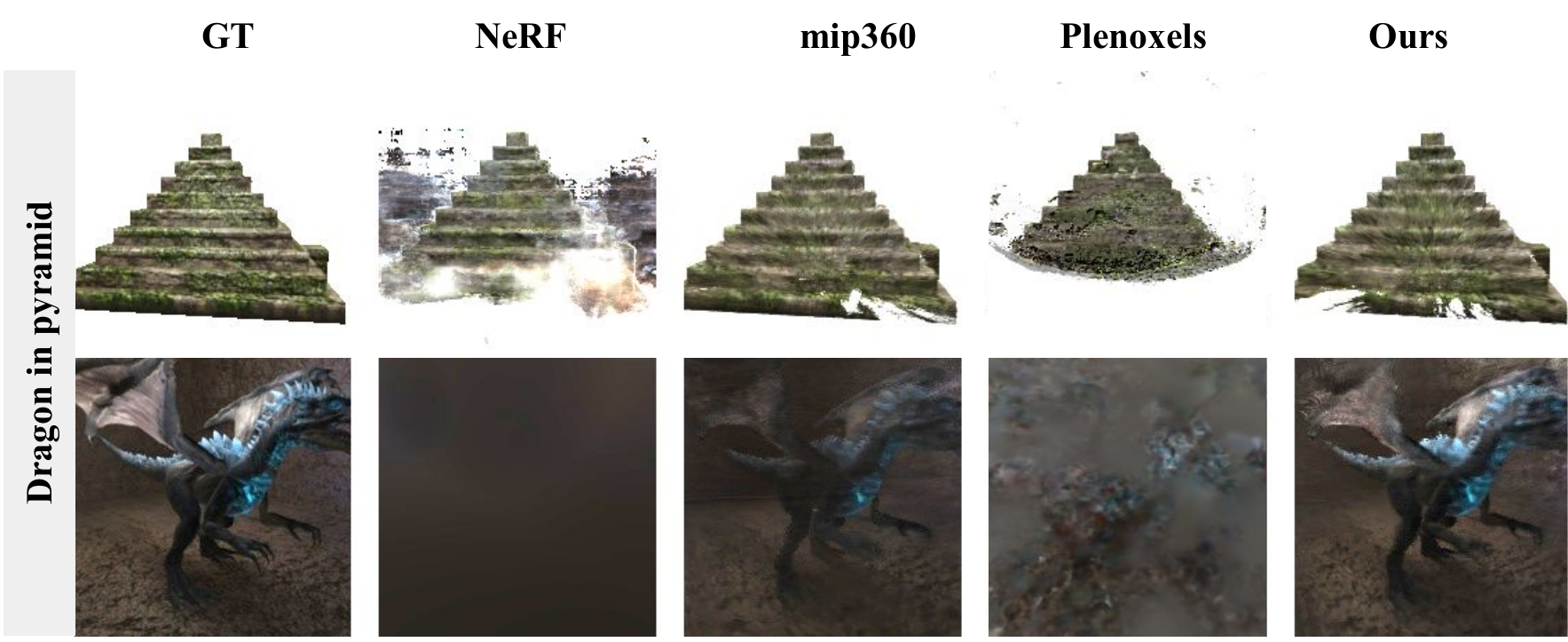}
    \caption{(From top to bottom) Qualitative results on the proposed synthetic datasets. Each row represents a novel view from each level of the stratified scene. The ground-truth view is shown in Column 1. Compared to prior works (Column 2-4) our method's (Column 5) renderings are more similar to the ground-truth.}
    \label{fig:qual_results_2}
\end{figure}

\begin{table}[!t]
\caption{Quantitative comparison of our model and mip360 baseline on Six Layer Scene.}
\label{tab:summary_real10}
\begin{adjustbox}{width=\linewidth}
\begin{tabular}{@{}c|c|cc|cc@{}}
\toprule
\textbf{Dataset} & \textbf{Levels} & \textbf{mip360~\cite{barron2022mip360}} & \textbf{Ours} & \textbf{mip360~\cite{barron2022mip360}} & \textbf{Ours} \\ \midrule
\textbf{\textit{Spanish Colonial Retreat}} & 5 & 20.106 & \textbf{22.514} & 0.622 & \textbf{0.685} \\
\textbf{\textit{31 Brian Dr Rochester}} & 4 & 23.273 & \textbf{28.026} & 0.715 & \textbf{0.835} \\
\textbf{\textit{139 Barton Avenue}} & 6 & 18.991 & \textbf{23.433} & 0.642 & \textbf{0.780} \\ 
\textbf{\textit{7 Rutledge Ave}} & 7 & 19.621 & \textbf{25.040} & 0.566 & \textbf{0.791} \\\bottomrule
\end{tabular}
\end{adjustbox}
\end{table}
\noindent \textbf{Qualitative Results.} Figure \ref{fig:qual_results_1} \& \ref{fig:qual_results_2}  depicts the qualitative results for the synthetic dataset scenes described in Section \ref{subsec:dataset}. We observe that NeRF~\cite{nerf} performs poorly regardless in majority of scenarios. The generated novel views for ``Coffee Shop" are poor. It only works well in level 0 of ``Cube-Sphere-Monkey" dataset. mip360~\cite{barron2022mip360}  outperforms NeRF but falters in level 1. Furthermore, in level 0 of the ``Cube-Sphere-Monkey" dataset, mip360 only generates a white patch with no visible structure. For RealEstate10K dataset, it can be observed in Figure~\ref{fig:qual_real10k} that mip360 generates blurry results compared to our approach. Further, we find that our approach generates consistent and structurally salient novel views throughout all levels and scenes. We show qualitative results for Instant-NGP~\cite{muller2022instant} and TensoRF~\cite{chen2022tensorf} in Appendix ~\refsupp{E.2} in the supplementary material.

\begin{figure}[!th]
    \centering
    \includegraphics[width=\linewidth]{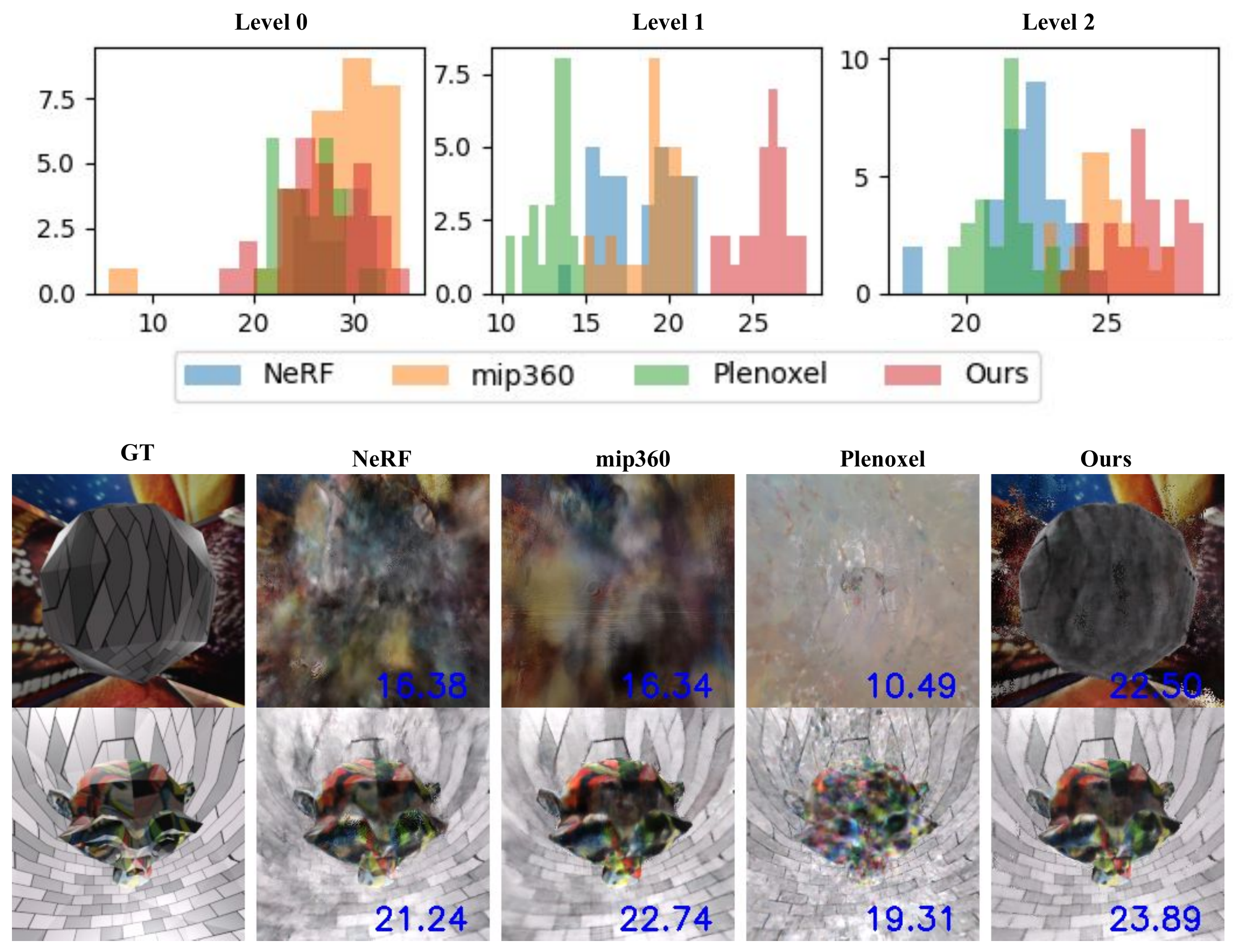}
    \caption{(Top Row) Comparison of histogram plots for the test-set for PSNR on \textbf{``Cube-Sphere-Monkey''}. Note how distribution of our our method is always towards the right compared to other methods. $x-axis$ denote metric value and $y-axis$ denotes the frequency. A qualitative comparison of our method's worst-case PSNR results. \textcolor{blue}{PSNR} is present at the bottom of the result image.}
    \vspace{-4mm}
    \label{fig:qualitative_worst_case}
\end{figure}

\noindent \textbf{Worst Case Analysis.} When comparing different methods, average metrics are often insufficient to determine which method is superior to the others. As we have observed in Figure \ref{fig:qualitative_worst_case} that the baseline method fails on some of test images, hence we also compare the methods in worst care scenarios. The worst-case analysis describes a method's worst performance on the dataset. The worst case analysis is particularly useful to detect the shortcomings of the methods. We present analysis in two categories: (a) histogram distribution for each metric on the test set, and (b) qualitative comparison of the worst-case scenario for our method on PSNR metric. 

Figure \ref{fig:qualitative_worst_case} compares PSNR histogram plots on test-set views for the \textbf{``Cube-Sphere-Monkey''} scene. We can see that the mip360 approach performs poorly on PSNR and ranks low on practically all stratification levels. This supports our argument that the mip360 approach produces artifacts in such stratified scenes. For our method, the PSNR distributions are on the right. This implies that the novel views on test-set from our method will not be having serious artifacts in most cases, demonstrating its reliability. 

\begin{figure*}[!t]
    \centering
    \includegraphics[width=\linewidth]{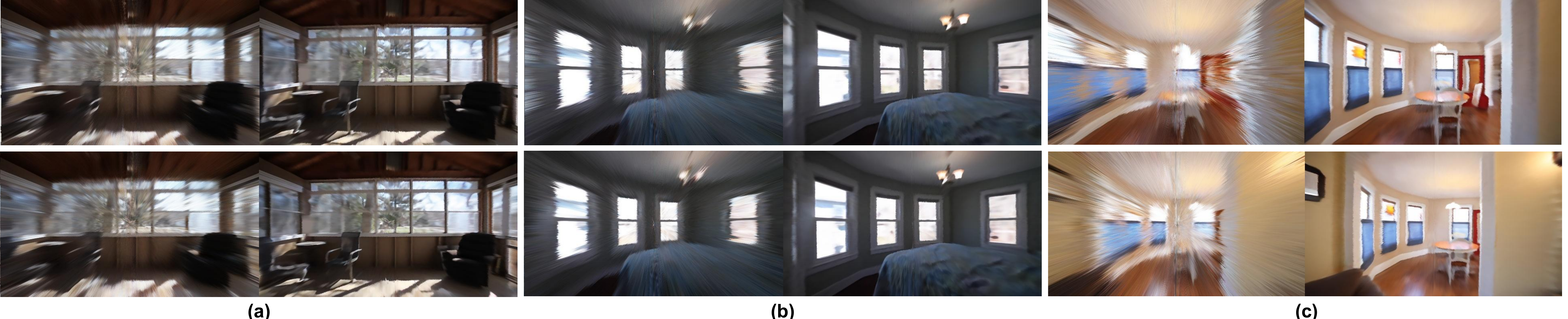}
    \caption{Novel-views from different levels of 'Real Estate Video Tour 7 Rutledge Ave Highland Mills NY 10930  Orange County NY' scene in Real Estate 10K dataset. The two rows are from two-different view-points.}
    \label{fig:qual_results_r10k_2}
\end{figure*}

Images in Figure \ref{fig:qualitative_worst_case} depict the qualitative results for the worst-case PSNR instances. All methods perform well in level 0. Hence, we are discussing interior levels which are level 1 and level 2. Other approaches fail in the worst-case scenario for our method at level 1. The outputs from NeRF, mip360 and Plenoxel are visually impaired. At level 2, our method has less blur compared to other approaches. These findings demonstrate that our method is better suited to represent stratified scenes than others.

\begin{table}[!t]
\caption{Quantitative comparison of our model and baseline on Synthetic Six Layer Scene.}
\label{tab:six_layer}
\begin{adjustbox}{width=\linewidth}
\begin{tabular}{@{}cccccccc@{}}
\toprule
 &
  \textbf{Metrics} &
  \textbf{Level 0} &
  \textbf{Level 1} &
  \textbf{Level 2} &
  \textbf{Level 3} &
  \textbf{Level 4} &
  \textbf{Level 5} \\ \midrule
\multicolumn{1}{c|}{\multirow{2}{*}{\textbf{mip360~\cite{barron2022mip360}}}} &
  \multicolumn{1}{c|}{\textbf{PSNR $\uparrow$}} &
  22.215 &
  16.183 &
  15.084 &
  12.012 &
  21.813 &
  21.539 \\
\multicolumn{1}{c|}{} &
  \multicolumn{1}{c|}{\textbf{SSIM $\uparrow$}} &
  0.777 &
  0.442 &
  0.510 &
  0.344 &
  0.817 &
  0.647 \\ \midrule
\multicolumn{1}{c|}{\multirow{2}{*}{\textbf{Ours}}} & \multicolumn{1}{c|}{\textbf{PSNR $\uparrow$}} & \textbf{23.889} & \textbf{21.449} & \textbf{21.456} & \textbf{24.095} & \textbf{28.283} & \textbf{21.898} \\
\multicolumn{1}{c|}{} &
  \multicolumn{1}{c|}{\textbf{SSIM $\uparrow$}} &
  \textbf{0.833} &
  \textbf{0.681} &
  \textbf{0.685} &
  \textbf{0.722} &
  \textbf{0.883} &
  \textbf{0.686} \\ \bottomrule
\end{tabular}
\end{adjustbox}
\end{table}

\begin{figure}[!t]
    \centering
    \includegraphics[width=\linewidth]{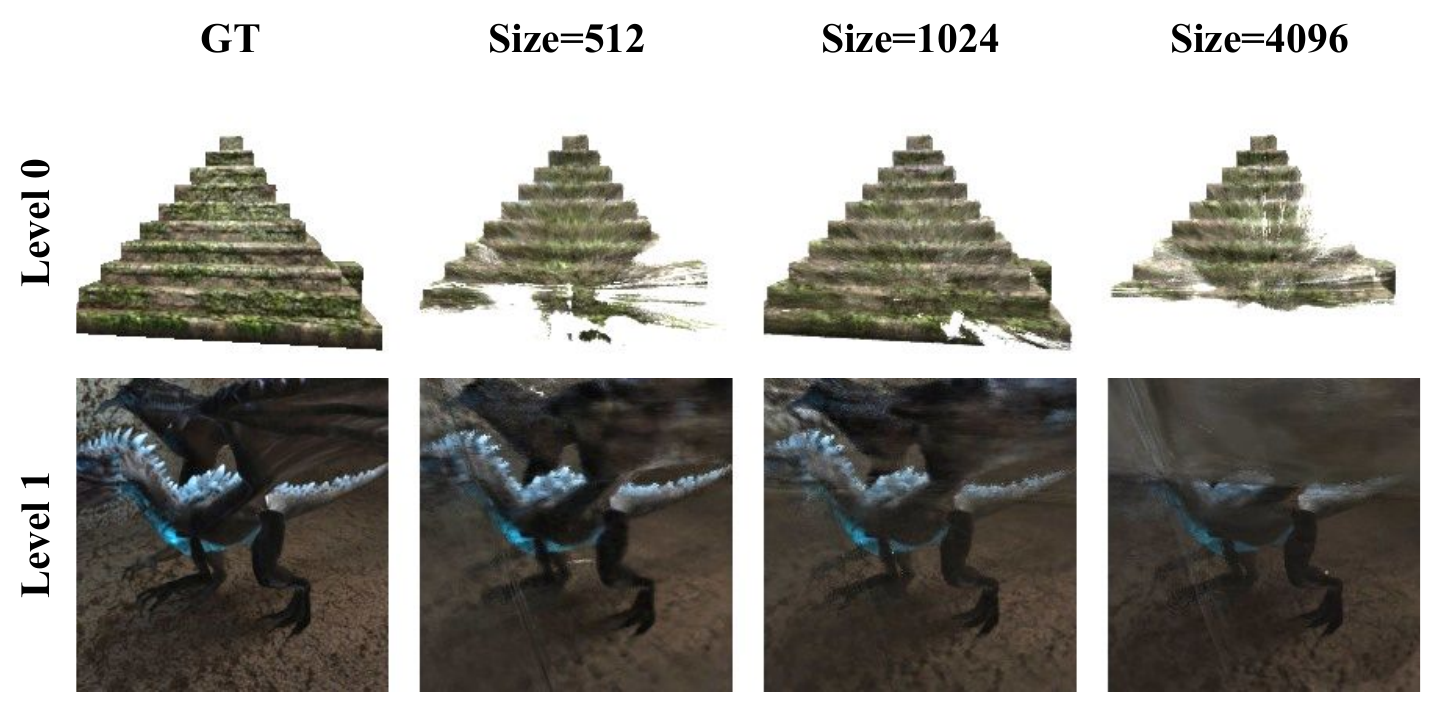}
    \caption{Comparisons of different codebook size on \textbf{``Dragon in Pyramid''} scene for different vector-codebook sizes. Note at size=1024 we achieve the best results with less artifacts.}
    \label{fig:ablation_codebook_size}
    \vspace{-4mm}
\end{figure}

\noindent \textbf{Ablation Studies.} To analyse our proposed method, we present an ablation on the size of the vector codebook in our latent generator. Table~\ref{tab:ablation_codebook_size}  shows the ablation for the size of the vector codebook on the ``Coffee Shop" dataset. We trialed with codebook sizes of $512$, $1024$ and $4096$. We found that size $1024$ provides optimal performance. As shown in Figure \ref{fig:ablation_codebook_size}, increasing the codebook size induces haziness in the generated novel views, while decreasing the size creates white artifacts in level 0. As a result, we fix the size $1024$ for all of our synthetic experiments. Whereas for RealEstate10K dataset we find that codebook size of $4096$ produces the optimal tradeoff of results across levels, as it contains more number of levels and details.We further discuss the key architectural design choices for Latent Generator and Latent Router modules in Appendix ~\refsupp{E.5}.

\noindent \textbf{No. of levels}: To further test the efficacy of our method on higher number of levels, we created a \textit{``Simple Geometry''} scene consisting of primitive geometry shapes like cube and spheres. More details are in the supplementary material. Table \ref{tab:six_layer} displays the results for both the baseline and our approach across a six levels stratified scene. The average PSNR/SSIM for the mip360 baseline is \textbf{15.35 / 0.487}, while our method achieved PSNR/SSIM of \textbf{23.54 / 0.754} which improves PSNR and SSIM by \textbf{53.35 \%} and \textbf{54.83 \%} respectively. This shows that our method performs better on increasing number of levels when compared with the baseline method. These observations also hold true for scenes in the RealEstate10K dataset as shown in Table~\ref{tab:six_layer_real10k}.

\begin{table}[!t]
\caption{Quantitative results on \textbf{``Cube-Sphere-Monkey"} scene for ablation on size of the vector codebook in Latent Generator.}
\label{tab:ablation_codebook_size}
\begin{adjustbox}{width=\linewidth}
\begin{tabular}{c|cc|cc|cc}
\hline
\textbf{Size} & \multicolumn{2}{c|}{\textbf{PSNR} $\uparrow$}  & \multicolumn{2}{c|}{\textbf{SSIM} $\uparrow$}  & \multicolumn{2}{c}{\textbf{LPIPS} $\downarrow$}  \\ \cline{2-7} 
              & \textbf{Level 0} & \textbf{Level 1} & \textbf{Level 0} & \textbf{Level 1} & \textbf{Level 0} & \textbf{Level 1} \\ \hline
 \textbf{512} &
  \textbf{29.5458} &
  {26.3497} &
  \textbf{0.8743} &
  {0.7395} &
  0.1675 &
  \textbf{0.4899} \\
\textbf{1024} &
  29.4834 &
  {26.1715} &
  0.8701 &
  \textbf{0.7489} &
  \textbf{0.1367} &
  0.5163 \\
\textbf{4096} & 28.4609 & \textbf{27.8274} & 0.8628 & 0.7342 & 0.1776           & 0.5027          \\ \hline
\end{tabular}
\end{adjustbox}
\end{table}

\section{Conclusion}
In this work, we focus on the problem of modelling the 3D representation of a stratified and hierarchical scene, implicitly through a single neural field. For this,  we propose \emph{Strata-NeRF}, which models scenes with stratified structures by introducing a VQ-VAE-based latent generator to implicitly learn the distribution of latent space of input 3D locations and condition the neural radiance field with the latent code generated from this distribution. We also introduce a new synthetic dataset with  stratified-level scenes and use it to analyse various existing approaches. Through quantitative, qualitative, and worst-case analysis on this dataset, we show that \emph{Strata-NeRF} has a more stable 3D representation than the other methods. Further, the improvements due to \emph{Strata-NeRF} also generalize to real-world RealEstate10K dataset, where it outperforms baselines by a significant margin establishing a new state-of-the-art. We believe designing a new volume rendering equation for modelling complex stratified scenes is a good direction for future work.

\vspace{1mm}

\noindent \textbf{Acknowledgement.} This work was supported by Samsung R\&D Institute India, Bangalore, PMRF and Kotak IISc AI-ML Centre (KIAC). Srinath Sridhar was partly supported by NSF grant CNS-2038897

\renewcommand \thepart{}
\renewcommand \partname{}

\doparttoc %
\faketableofcontents %

\appendix
\renewcommand*\contentsname{Appendix}
\addcontentsline{toc}{section}{Appendix} %
\part{Appendix} %
\parttoc %

\begin{figure*}[!htb]
    \centering
    \includegraphics[width=\linewidth]{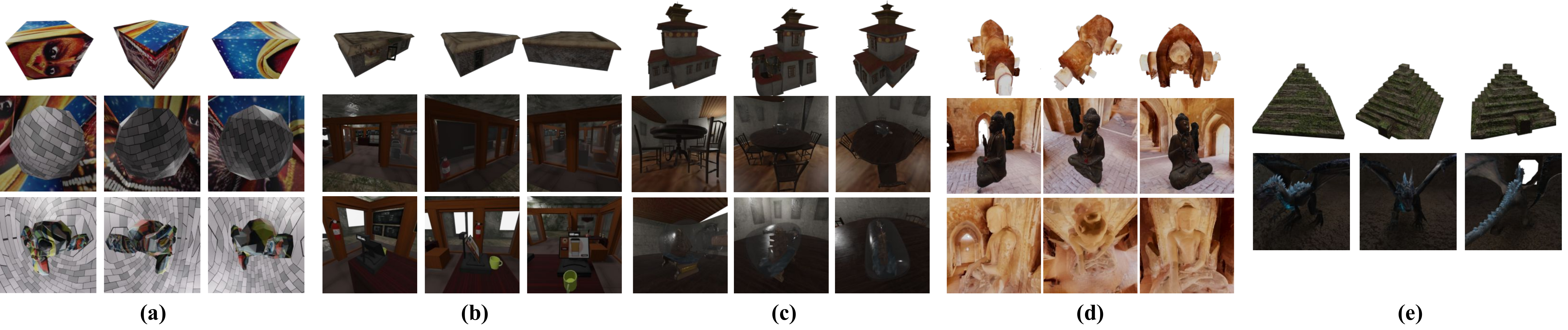}
    \caption{(a) Cube-Sphere-Monkey, (b) Coffee Shop, (c) Bhutanese House, (d) Buddhist Temple and (e) Dragon In Pyramid. Representative images for each level. }
    \label{fig:scene_representation}
\end{figure*}
\section{Introduction}
We present additional results and other details related to our proposed method : Strata-NeRF. We elaborate on the proposed synthetic stratified dataset in Appendix \ref{sec:dataset_details}. We give the implemntation details in Appendix \ref{sec:implementation_details}. Then, we present additional ablation study and results in Appendix \ref{sec:additional_experiments}.

\section{Synthetic Dataset Details}
\label{sec:dataset_details}
Figure \ref{fig:scene_representation} shows the representation of each level of each scene. Table \ref{tab:num_views_table} shows the level-wise split for each scene. 

\subsection{Cube-Sphere-Monkey}
This dataset consists of simple geometric entities such as a cube, sphere and a monkey mesh provided in Blender~\cite{blender}. Figure \ref{fig:scene_representation} illustrates the layout of this scene. \textit{Cube} is at level 0, \textit{Sphere} is at level 1 and \textit{Monkey} is at the innermost level. The texture for \textit{Cube} is an image generated from Stable Diffusion demo~\cite{huggingfaceSD}. We sample camera poses from the curved surface of a hemisphere for the outer cube and from the curved surface of a sphere for the inner levels.

\subsection{Coffee Shop}
This dataset mimics an actual coffee shop setup inside another shopping complex. The outermost level consists of concrete walls. At level 1, i.e. when one enters the shopping complex, there is regular flooring and a concrete ceiling. Here, we also notice the exterior walls of our coffee shop. At level 2; i.e., inside the coffee shop; there is a layout with a counter, menu board and a table for visitors. All these scenes are composited with the help of Blender~\cite{blender}. We sample camera poses from the curved surface of a hemisphere for all the levels.

\subsection{Bhutanese House}
A typical household setting inspired us to create this dataset. A typical residence features a table in the living room. In most cases, a decorative object is kept on the table. For the structure of the house, we choose a Bhutanese house model. The exterior of this structure is level 0. At level 1, i.e., inside the house, there are chairs, tables and other household items in the living room. At level 2, we have a glass bottle with a ship. We sample camera poses from the curved surface of a hemisphere. For level 2, we capture around the glass bottle on the circular table.

\begin{table}[!t]
\caption{\textbf{train-val-test} level-wise split for each scene.}
\label{tab:num_views_table}
\begin{adjustbox}{width=\linewidth}
\begin{tabular}{@{}c|c|ccc@{}}
\toprule
\textbf{Scene} & \textbf{Split} & \textbf{Level 0} & \textbf{Level 1} & \textbf{Level 2} \\ \midrule
\multirow{3}{*}{\textbf{Cube-Sphere-Monkey}} & \textbf{train} & 30 & 30 & 30 \\
                                    & \textbf{val}   & 30 & 30 & 30 \\
                                    & \textbf{test}  & 30 & 30 & 30 \\ \midrule
\multirow{3}{*}{\textbf{Coffee Shop}} & \textbf{train} & 30 & 30 & 30 \\
                                    & \textbf{val}   & 15 & 15 & 15 \\
                                    & \textbf{test}  & 15 & 15 & 15 \\ \midrule
\multirow{3}{*}{\textbf{Bhutanese House}} & \textbf{train} & 30 & 30 &30  \\
                                    & \textbf{val}   & 15 & 15 & 15 \\
                                    & \textbf{test}  & 15 & 15 & 15 \\ \midrule
\multirow{3}{*}{\textbf{Buddhist Temple}} & \textbf{train} & 30 & 20 &20  \\
                                    & \textbf{val}   & 15 & 10 & 10 \\
                                    & \textbf{test}  & 15 & 10 & 10 \\ \midrule
\multirow{3}{*}{\textbf{Dragon In Pyramid}} & \textbf{train} & 30 & 30  & - \\
                                    & \textbf{val}   & 15 & 15 & - \\
                                    & \textbf{test}  & 15 & 15 & - \\ \bottomrule
\end{tabular}
\end{adjustbox}
\end{table}

\subsection{Dragon In Pyramid}
This dataset captures a fantastical world filled with pyramids and dragons. We use a model of a \textit{Mayan pyramid} as the outer structure. Inside the pyramid, we place a model of a dragon. Thus, this scene has two levels: 1.) the outer walls of the \textit{Mayan pyramid} and 2.) the dragon residing inside the pyramid. All the camera poses are sampled from the curved surface of different hemispheres. 

\subsection{Buddhist Temple}
This scene depicts an archaeological site or a typical monument location. We select a Buddhist temple to represent this scene. Two levels indicate the nearby rooms inside the structure in this context. Level 0 represents the outer structure of the monument, Levels 1 contains a bronze statue in the center of the monument, and Level 2 contains a Buddha statue mounted to the wall of one room.

\section{Real Dataset}
\label{sec:real_data_set}
We evaluate our method on real-world scenes as well. We choose RealEstate10K~\cite{zhou2018stereo} dataset, which contains camera poses corresponding to camera frames from video-clips exracted from Youtube videos. The camera poses are obtained  by running SLAM and bundle adjustment algorithm over these large videos. To create a ``stratified'' scene from this dataset, first we cluster video clips belonging to same Youtube video using the video token provided in the ground-truth files. Then we extracted camera frames and pose as per the timestamp information provided in the ground-truth files. The extracted camera pose for each video clip from a scene were already aligned with respect to a common coordinate system. We removed the video clips which had any dynamic motion within them. We extracted four scenes which are ``Spanish Colonial Retreat in Scottsdale Arizona''~\cite{scene0}, ``139 Barton Avenue Toronto Ontario''~\cite{scene50} ,``31 Brian Dr Rochester NY''~\cite{scene60} and ``7 Rutledge Ave Highland Mills''~\cite{scene72}.   

\section{Implementation Details}
\label{sec:implementation_details}
\begin{figure}[!b]
    \centering
    \includegraphics[width=\linewidth]{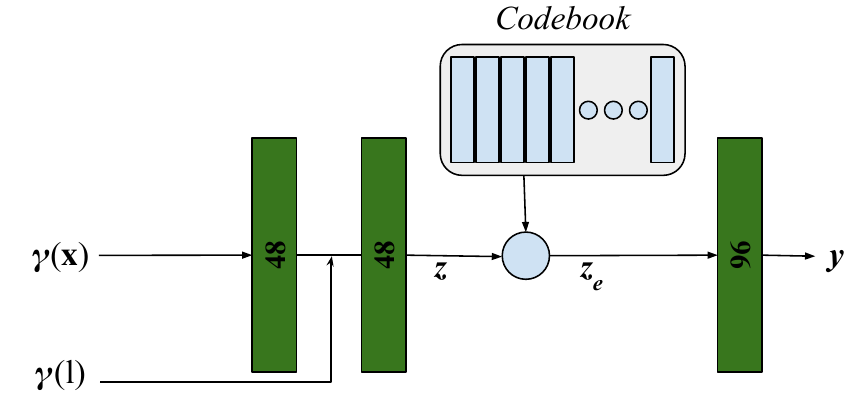}
    \caption{A diagram of ``Latent Generator'' network. This network takes position-encoded 3D point $\gamma(x)$ and position-encoded camera level $\gamma(l)$. This is passed through the encoder block to get $z$ which is than matched to the nearest latent in the codebook to get $z_e$. $z_e$ is passed through decoder block to reconstruct the position-encoded 3D point $y$. }
    \label{fig:latent_gen_network}
\end{figure}

\begin{figure}[!b]
    \centering
    \includegraphics[width=\linewidth]{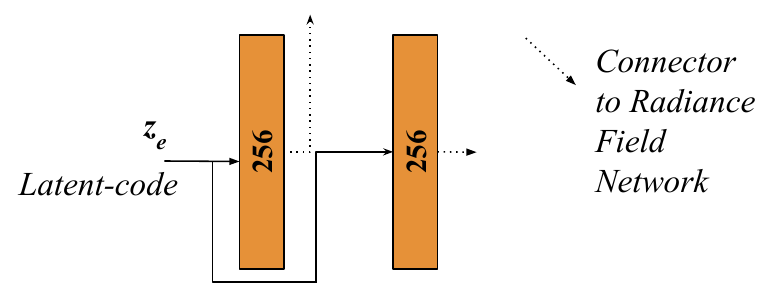}
    \caption{A diagram of ``Latent Router'' network. This network takes latent code $z_e$ generated by the ``Latent Generator'' and connects it to the radiance field network after passing through linear layers.}
    \label{fig:latent_router_network}
\end{figure}

\textbf{Architecture Details.} We provide architectural details of the ``Latent Generator'' and ``Latent Router'' networks in Figure \ref{fig:latent_gen_network} and \ref{fig:latent_router_network} respectively.

\textbf{Training.} We use Adam~\cite{kingma2014adam} optimizer with hyperparameters $\beta_1 = 0.9$, $\beta_2=0.999$, $\epsilon=1e^{-6}$ and initial learning rate $=0.002$. Further, the learning rate is log-linearly interpolated such that learning rate $=0.00002$ at maximum steps. Additionaly, there are $512$ warmup steps. Distortion loss proposed in Mip-NeRF 360~\cite{barron2022mip360} is switched off for the blender datasets as proposed by the authors. We use one proposal MLP and one NeRF MLP. We weight the loss for ``Latent Generator'' with value $\lambda_2 = 0.1$.

\textbf{Implementation.} Our implementation is based on Mip-NerF 360~\cite{barron2022mip360}  which uses JAX~\cite{bradbury2018jax} framework. 

\subsection{Choice of Training Configuration File}
The dataset described in Section \ref{sec:dataset_details} is created using Blender~\cite{blender}. This dataset has white background for the level 0. Barron \textit{et al.}~\cite{barron2022mip360} uses ``\textit{\text{blender\_256.gin}}'' file for the blender scenes proposed in NeRF~\cite{nerf} which are small in size compared to our scenes. This configuration file does not work for the scenes we proposed in Appendix \ref{sec:dataset_details}. Hence, we use ``\textit{\text{360.gin}}'' and alter the dataset type field in the configuration file.

\begin{figure}[!t]
    \centering
    \includegraphics[width=\linewidth]{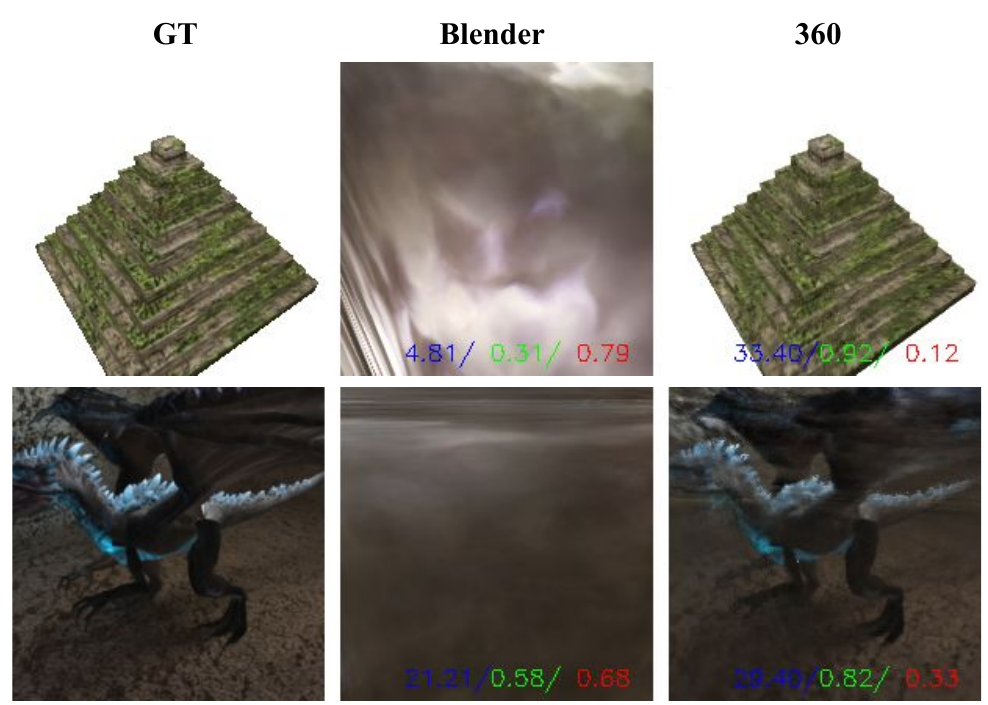}
    \caption{Qualitative comparison for different configuration files on \textit{Dragon In Pyramid} scene. We observe that \textit{360.gin} configuration generates better results. Metrics \textcolor{blue}{PSNR},\textcolor{green}{SSIM} and \textcolor{red}{LPIPS} are color-coded at the bottom of the result image}
    \label{fig:qualitative_diff_configs}
\end{figure}

Table \ref{tab:config_comparison} shows the quantitative comparison of the above mentioned configuration files on \textit{Dragon In Pyramid} dataset. We observe that the ``\textit{\text{360.gin}}'' configuration beats the  ``\textit{\text{blender\_256.gin}}'' in all the levels. Figure \ref{fig:qualitative_diff_configs} compares the qualitative results of these two configuration files. We notice that the novel views from ``\textit{\text{blender\_256.gin}}' are inferior in quality compared to ``\textit{\text{360.gin}}'' configuration.  ``\textit{\text{360.gin}}'' configuration  has better performance because of the contract function proposed by Barron~\cite{barron2022mip360}. The contract function is defined as follows:
\begin{equation}
    \mathrm{contract}(x) = \left\{ \begin{array}{ll} x, &  {\left||x||\right| \le {1}} \\ (\mathrm{{2} - \frac{1}{||x||}})\left(\frac{x}{||x||}\right), & \mathrm{otherwise} \end{array} \right.
\end{equation}
This contract function maps input coordinates onto a ball of radius $2$. Effectively, a large range is bounded inside a radius of $2\,m$. This is the reason why ``\textit{\text{360.gin}}'' configuration is better for large blender scenes. Hence, we use this configuration file for all the scenes other than ``Cube-Sphere-Monkey''.

\begin{table}[!t]
\caption{Performance on the \textit{Dragon In Pyramid} dataset between two configuration files. We observe that ``\textit{360.gin}'' works much better than the other configuration file.}
\label{tab:config_comparison}
\begin{adjustbox}{width=\linewidth}
\begin{tabular}{@{}c|ccc|ccc|ccc@{}}
\toprule
\multirow{2}{*}{\textbf{Config}} &
  \multicolumn{3}{c|}{\textbf{Level 0}} &
  \multicolumn{3}{c|}{\textbf{Level 1}} &
  \multicolumn{3}{c}{\textbf{Total}} \\ \cmidrule(l){2-10} 
 &
  \textbf{PSNR} &
  \textbf{SSIM} &
  \textbf{LPIPS} &
  \textbf{PSNR} &
  \textbf{SSIM} &
  \textbf{LPIPS} &
  \textbf{PSNR} &
  \textbf{SSIM} &
  \textbf{LPIPS} \\ \midrule
\textbf{Blender} & {5.5654} & {0.3717} & {0.6252} & {22.9489} & {0.6320} & {0.5844} & {14.2571} & {0.5018} & {0.6048}  \\
\textbf{360}    & \textbf{30.8758}  & \textbf{0.9006}   & \textbf{0.1367}   & \textbf{24.3890}   & \textbf{0.7054}   & \textbf{0.5163}   & \textbf{27.6324}   & \textbf{0.8030}   & \textbf{0.3265}  \\ \bottomrule
\end{tabular}
\end{adjustbox}
\end{table}

\section{Additional Experiments}
\label{sec:additional_experiments}
\begin{table}[!t]
\caption{No. of training parameters \textbf{(in millions)} for level-wise mip360 and our method with two different codebook sizes 1024 and 4096 for different number of levels.}
\label{tab:param_analysis}
\begin{adjustbox}{width=\linewidth}
\begin{tabular}{@{}cccc@{}}
\toprule
Levels &
  \begin{tabular}[c]{@{}c@{}}Level-Wise\\ mip360\end{tabular} &
  \begin{tabular}[c]{@{}c@{}}Ours\\ (1024 codebook)\end{tabular} &
  \begin{tabular}[c]{@{}c@{}}Ours\\ (4096 codebook)\end{tabular} \\ \midrule
1 & 0.835 & 0.924 & 1.071 \\
3 & 2.506 &   0.924    &   1.071    \\
4 & 3.341 &   0.924    &   1.071    \\
5 & 4.176 &  0.924     &   1.071    \\
6 & 5.011 &   0.924    &    1.071   \\ \bottomrule
\end{tabular}
\end{adjustbox}
\end{table}

\begin{figure}[!t]
    \centering
    \includegraphics[width=\linewidth]{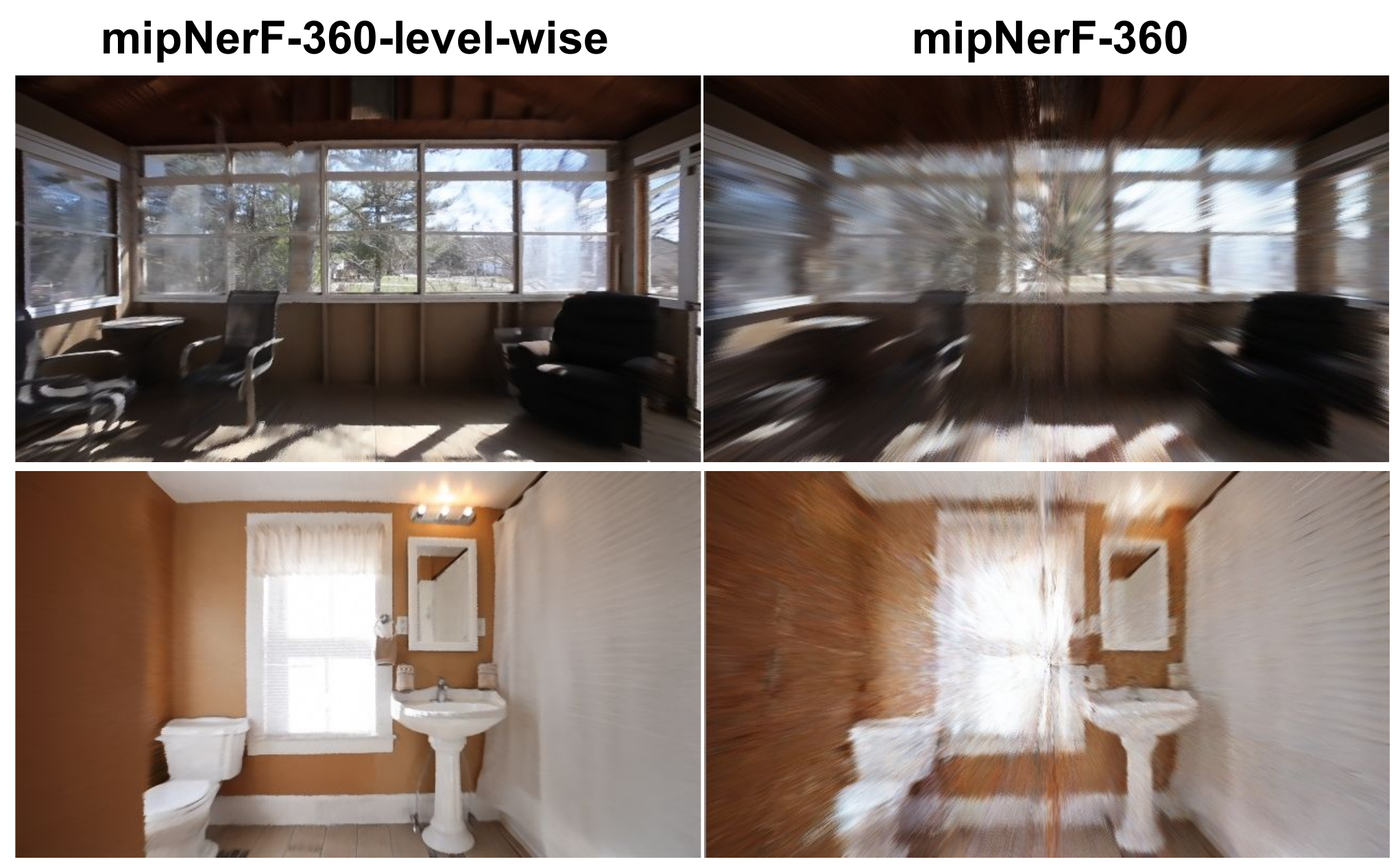}
    \caption{Analysis on “7 Rutledge Ave” scene from RealEstate10K ~\cite{zhou2018stereo} dataset. We present visual results from two levels. Note how artifacts appear in results from mipNeRF-360 (all levels are trained jointly) whereas when mipNeRF-360 is used for each level separately (level-wise) we observe no artifacts.}
    \label{fig:motivation_realestate}
\end{figure}

\begin{table}[!t]
\caption{A quantitative comparison of mip360 (level-wise) and mipNeRF-360 (all views) on ``7 Rutledge Ave''}
\label{tab:motivation_realestate}
\begin{adjustbox}{width=\linewidth}
\begin{tabular}{@{}ccccccccc@{}}
\toprule
\multicolumn{1}{c|}{\textbf{Methods}} & \textbf{Lv 0}  & \textbf{Lv 1}  & \textbf{Lv 2}  & \textbf{Lv 3}  & \textbf{Lv 4}  & \textbf{Lv 5}  & \textbf{Lv 6}  & \textbf{Total} \\ \midrule
\multicolumn{1}{c|}{\textbf{mipNeRF-360 (x7)}}  & \textbf{24.20} & 22.42 & \textbf{26.72} & 24.78 & 22.73 & \textbf{27.41} & 24.78 & \underline{24.25} \\
\multicolumn{1}{c|}{\textbf{mipNeRF-360}}       & 19.53 & 18.33 & 23.52 & 17.00 & 18.82 & 19.73 & 21.60 & 19.62 \\ \bottomrule
\end{tabular}
\end{adjustbox}
\end{table}
\subsection{RealEstate10K~\cite{zhou2018stereo} scene - Motivation Experiment}
We presented motivation of our work on a synthetic scene ``Dragon In Pyramid'' in Section \textcolor{red}{4} in the main paper. We observed that no artifacts are observed if individual mipNeRF-360 is trained for each level (level-wise) separately. We performed a similar experiment on the RealEstate10k~\cite{zhou2018stereo} scene and observed artifact-free novel views from level-wise mipNeRF-360. Similar to the observation for synthetic scenes, if all levels are trained combinedly we observe the artifacts in the rendered novel-views as shown in Fig ~\ref{fig:motivation_realestate}. Further, PSNR values in Tab. ~\ref{tab:motivation_realestate} for level-wise mipNeRF-360, with 7 radiance fields (x7) are higher compared to a single mipNeRF-360 for all-levels. This further substantiates our claim that a single mipNeRF-360 network is not able to learn all the stratified levels. 

\subsection{Comparison with InstantNGP~\cite{muller2022instant} and TensoRF~\cite{chen2022tensorf}}
\begin{figure*}[!t]
    \centering
    \includegraphics[width=\linewidth]{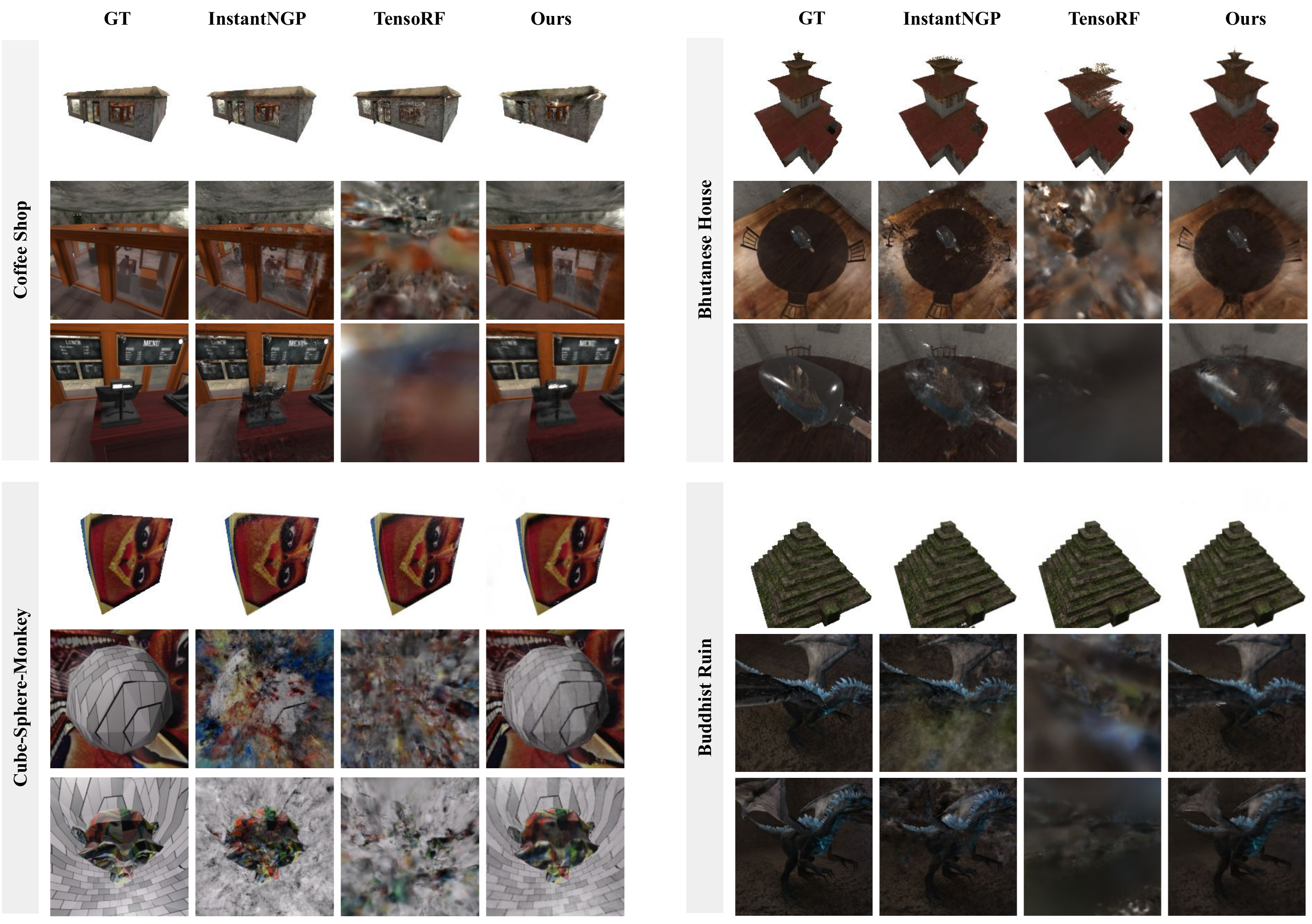}
    \caption{Qualitative Comparison on synthetic dataset for InstantNGP~\cite{muller2022instant} and TensoRF~\cite{chen2022tensorf} }
    \label{fig:qual_syn_ngp_1}
\end{figure*}

\begin{figure*}[!t]
    \centering
    \includegraphics[width=\linewidth]{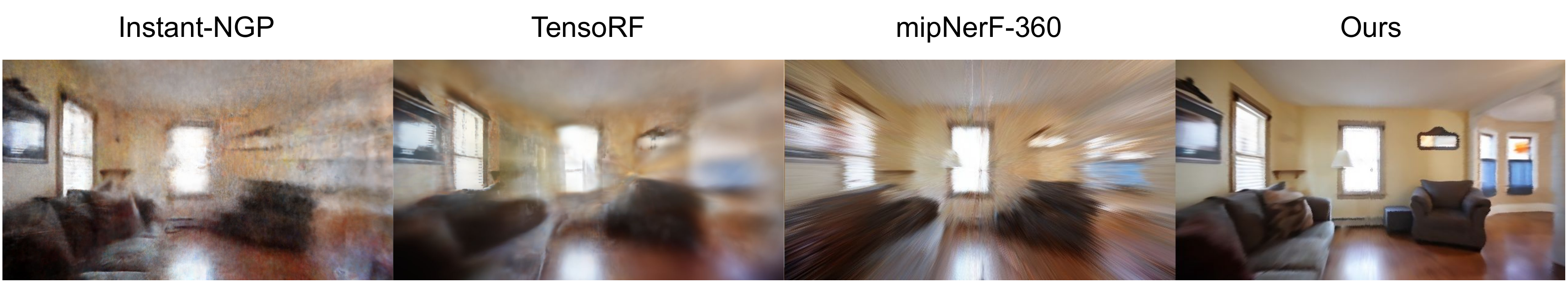}
    \caption{Qualitative Comparison on “7 Rutledge Ave” scene from RealEstate10K ~\cite{zhou2018stereo} dataset. The novel-view generated from our method is better than InstantNGP~\cite{muller2022instant}, TensoRF~\cite{chen2022tensorf} and mipNeRF-360~\cite{barron2022mip360}}
    \label{fig:qual_real10k_ngp}
\end{figure*}

\textbf{Synthetic Scenes. }We present qualitative comparison with InstantNGP~\cite{muller2022instant} and TensoRF~\cite{chen2022tensorf} in Fig. ~\ref{fig:qual_syn_ngp_1} and ~\ref{fig:qual_syn_ngp_2}. These methods work well in the outermost level. But suffer from artifacts because of the stratified scenes in the inner levels. We observe this pattern consistenly across all the synthetic scenes.

\begin{table}[!t]
\caption{A quantitative comparison of InstantNGP~\cite{muller2022instant} and TensoRF~\cite{chen2022tensorf} on ``7 Rutledge Ave''}
\label{tab:comparison_ngp_quant}
\begin{adjustbox}{width=\linewidth}
\begin{tabular}{@{}ccccccccc@{}}
\toprule
\multicolumn{1}{c|}{\textbf{Methods}} & \textbf{Lv 0}  & \textbf{Lv 1}  & \textbf{Lv 2}  & \textbf{Lv 3}  & \textbf{Lv 4}  & \textbf{Lv 5}  & \textbf{Lv 6}  & \textbf{Total} \\ \midrule

\multicolumn{1}{c|}{\textbf{Instant-NGP}}       & 19.02 & 18.24 & 21,32 & 19.43 & 18.77 & 18.98 & 21.33 & 19.47 \\
\multicolumn{1}{c|}{\textbf{TensoRF}}          & 18.03 & 21.29 & 21.23 & 20.23 & 20.36 & 18.57 & 22.69 & 20.70 \\ 
\multicolumn{1}{c|}{\textbf{Ours}}    & \textbf{22.84} & \textbf{25.14} & \textbf{24.83} & \textbf{25.67} & \textbf{25.15} & \textbf{23.10} & \textbf{26.75} & \textbf{25.04} \\ \bottomrule
\end{tabular}
\end{adjustbox}
\end{table}

\textbf{RealEsate10K~\cite{zhou2018stereo} dataset} Fig. ~\ref{fig:qual_real10k_ngp} shows qualitative comparison on “7 Rutledge Ave” scene from RealEstate10K ~\cite{zhou2018stereo}. Our method generates novel-view without any artifact, whereas other methods have visible artifacts in the generated novel-views. Tab. ~\ref{tab:comparison_ngp_quant} shows PSNR of the generated novel-views. Our method clearly outperforms InstantNGP~\cite{muller2022instant} and TensoRF~\cite{chen2022tensorf}. 

\subsection{Comparison with level-wise radiance fields.} One trivial solution for the proposed stratified setting is training mip360 individually for multi-view images in each level. We show that with increase in no. of levels, no. of training parameters increases linearly. Consider a mip360 network with width $256$ and depth $8$. We present varaition of no. of training parameters in Table \ref{tab:param_analysis} for different number of levels. Our method's training parameter requirement doesnot increase linearly as it does in level-wise mip360. 

For comparison, on ``Spanish Colonial Retreat'' scene, mipNerf-360 takes \textit{5h 30m} to train, while our method, with a vector-codebook size of 1024, takes \textit{6h 20m} for 150k iterations on a single NVIDIA RTX 3090 GPU.

\subsection{Ablation on Vector-Codebook Size}
We present more results on \textit{Coffee Shop, Bhutanese House and Buddhist Temple}  for the ablation : \textit{Size of the vector-codebook in ``Latent Generator''}. We tried with three sizes : 512, 1024 and 4096. Table \ref{tab:extra_ablation_size_codebook_1} %
and \ref{tab:extra_ablation_size_codebook_3}  shows the quantitative results for the mentioned datasets. We observe that vector codebook of size $1024$. gives us the overall best results.

\begin{table}[!t]
\caption{Performance on the \textit{Coffee Shop} dataset for different sizes of the vector codebook. \textbf{Best} results are marked in bold and \underline{Second-best} results are underlined.}
\label{tab:extra_ablation_size_codebook_1}
\begin{adjustbox}{width=\linewidth}
\begin{tabular}{@{}c|ccc|ccc|ccc|ccc@{}}
\toprule
\multirow{2}{*}{\textbf{Size}} &
  \multicolumn{3}{c|}{\textbf{Level 0}} &
  \multicolumn{3}{c|}{\textbf{Level 1}} &
  \multicolumn{3}{c}{\textbf{Level 2}} &
  \multicolumn{3}{c}{\textbf{Total}}\\ \cmidrule(l){2-13} 
 &
  \textbf{PSNR} &
  \textbf{SSIM} &
  \textbf{LPIPS} &
  \textbf{PSNR} &
  \textbf{SSIM} &
  \textbf{LPIPS} &
  \textbf{PSNR} &
  \textbf{SSIM} &
  \textbf{LPIPS} &
  \textbf{PSNR} &
  \textbf{SSIM} &
  \textbf{LPIPS}\\ \midrule
\textbf{512} &                    24.4768 &                    0.8605 &                    0.2049 &                    28.0758 &                    0.8257 &                    0.3632 & \textbf{33.7944}   & \underline{0.9306} & \textbf{0.2003}   &                    28.7824 &                    0.8723 &                    0.2561  \\
\textbf{1024} & \textbf{26.4497}   & \textbf{0.8803}   & \textbf{0.1936}   & \textbf{28.6387}   & \textbf{0.8403}   & \textbf{0.3449}   & 33.2695                    & 0.9254                    & 0.2243                    & \textbf{29.4526}   & \textbf{0.8820}   & \underline{0.2543} \\
\textbf{4096}    & \underline{25.3534} & \underline{0.8729} & \underline{0.1995} & \underline{28.4341} & \underline{0.8383} & \underline{0.3539} & \underline{33.6062} & \textbf{0.9316}   & \underline{0.2025} & \underline{29.1312} & \underline{0.8809} & \textbf{0.2520} \\ \bottomrule
\end{tabular}
\end{adjustbox}
\end{table}

\begin{table}[!t]
\caption{Performance on the \textit{Buddhist Temple} dataset for different sizes of the vector codebook. \textbf{Best} results are marked in bold and \underline{Second-best} results are underlined.}
\label{tab:extra_ablation_size_codebook_3}
\begin{adjustbox}{width=\linewidth}
\begin{tabular}{@{}c|ccc|ccc|ccc|ccc@{}}
\toprule
\multirow{2}{*}{\textbf{Size}} &
  \multicolumn{3}{c|}{\textbf{Level 0}} &
  \multicolumn{3}{c|}{\textbf{Level 1}} &
  \multicolumn{3}{c}{\textbf{Level 2}} &
  \multicolumn{3}{c}{\textbf{Total}}\\ \cmidrule(l){2-13} 
 &
  \textbf{PSNR} &
  \textbf{SSIM} &
  \textbf{LPIPS} &
  \textbf{PSNR} &
  \textbf{SSIM} &
  \textbf{LPIPS} &
  \textbf{PSNR} &
  \textbf{SSIM} &
  \textbf{LPIPS} &
  \textbf{PSNR} &
  \textbf{SSIM} &
  \textbf{LPIPS}\\ \midrule
\textbf{512} & \underline{27.3121} & \underline{0.8881} & \underline{0.1861} &                    25.3407 &                    0.7619 &                     0.362 & \underline{25.4983} & \underline{0.7476} & \underline{0.3691} & \underline{26.2306} & \underline{0.8119} & \underline{0.2886}  \\
\textbf{1024} & \textbf{27.5529}   & \textbf{0.8935}   & \textbf{0.1775}   & \textbf{27.3453}   & \textbf{0.7894}   & \textbf{0.3240}   & \textbf{25.5956}   & \textbf{0.7717}   & \textbf{0.3456}   & \textbf{26.9343}   & \textbf{0.8289}   & \textbf{0.2674} \\
\textbf{4096}    & 20.9017                    & 0.8075                    & 0.2680                    & \underline{27.0011} & \underline{0.7856} & \underline{0.3340} & 23.4656                    & 0.7189                    & 0.3853                    & 23.3769                    & 0.7759                    & 0.3204 \\ \bottomrule
\end{tabular}
\end{adjustbox}
\end{table}

\subsection{Architectural Design Choices.}
\label{subsec:arch_choices}
The proposed method consists of Latent Generator (LG) and Latent Router(LR) as shown in Figure \textcolor{red}{4} in the main paper. Latent Generator(LG) and Latent Router(LR) are described in Section \textcolor{red}{5.1} and \textcolor{red}{5.2} respectively in the main paper. To further motivate this choice of the architecture, we discuss the following design choices for the proposed method:
\begin{enumerate}
    \item Disabling the second router in \textbf{LR}: \textbf{D1}
    \item Disabling the first router in \textbf{LR}: \textbf{D2}
    \item removing \textbf{LR} and directly concatenating the generated embedding to the input positional encoding : \textbf{D3}
    \item Replacing the VQ-VAE block with the VAE block in \textbf{LG} : \textbf{D4}
\end{enumerate}
We present overall results for synthetic and RealEstate10K scenes in Tab. ~\ref{tab:arch_choices}. We conclude that using two parallel dense layers is better than an individual dense layer in \textbf{LR}. Further, we observe that how using Latent Router is better than directly concatenating the generated embedding with the input positional embedding. Similarly, the VAE version of our method underperforms the discrete VQ-VAE used in our method. 

\begin{table}[!t]
\caption{Ablation studies on the key design choices for the proposed method. \textbf{D1}: Disable second router in LR, \textbf{D2}: Disable first router in LR, \textbf{D3}: Remove LR and directly concatenate generated embedding with the positional encoding and \textbf{D4}: Replace VQ-VAE with VAE in LG. Acronyms D1, D2, D3, D4 are explained in more detail in Appendix ~\ref{subsec:arch_choices}}
\begin{adjustbox}{width=\linewidth}
\begin{tabular}{@{}cccccc@{}}
\toprule
\multicolumn{1}{c|}{}          & \textbf{D1}    & \textbf{D2}    & \textbf{D3}    & \textbf{D4}    & \textbf{Ours}  \\ \midrule
\multicolumn{1}{c|}{\textbf{Synthetic}}     & 26.04 & 27.34 & 27.41 & 26.96 & \textbf{28.25} \\
\multicolumn{1}{c|}{\textbf{RealEstate10K}} & 23.79 & 24.24 & 23.79 & 20.99 & \textbf{24.75} \\ \bottomrule
\end{tabular}
\label{tab:arch_choices}
\end{adjustbox}
\end{table}

\subsection{Why shared codebooks are important?}
We provide another ablation by creating independent code-book vectors for different levels : ``Ours-Ind.''. In our method, codebooks are shared between level which yield better results. This is natural as walls, etc. are shared between levels in the scene. 

\begin{table}[!t]
\caption{\textbf{Quantitative Comparison on ``7 Rutledge Ave''}}
\begin{adjustbox}{width=\linewidth}
\begin{tabular}{@{}ccccccccc@{}}
\toprule
\multicolumn{1}{c|}{\textbf{Ours-Ind.}}       & 21.03 & \underline{23.54} & 24.15 & \underline{23.85} & \underline{22.83} & 22.64 & \underline{25.41} & 23.53 \\
\multicolumn{1}{c|}{\textbf{Ours}}    & \underline{22.84} & \textbf{25.14} & \underline{24.83} & \textbf{25.67} & \textbf{25.15} & \underline{23.10} & \textbf{26.75} & \textbf{25.04} \\ \bottomrule
\end{tabular}
\end{adjustbox}
\end{table}

\subsection{Experiments on the standard novel-view synthesis dataset.}
We train the ``garden'' scene from the mipNeRF-360 dataset by treating it as a single-level scene. We achieved a PSNR of 26.40 on the test dataset, while mipNeRF-360 reports a PSNR of 26.98.  We achieve an average PSNR of 33.21 across all NeRF-synthetic scenes, while mipNeRF-360 achieves 33.09. Our proposed method performs comparably on these datasets, despite being designed for stratified scenes.

\subsection{Number of Views}
\begin{figure}[!t]
    \centering
    \includegraphics[width=\linewidth]{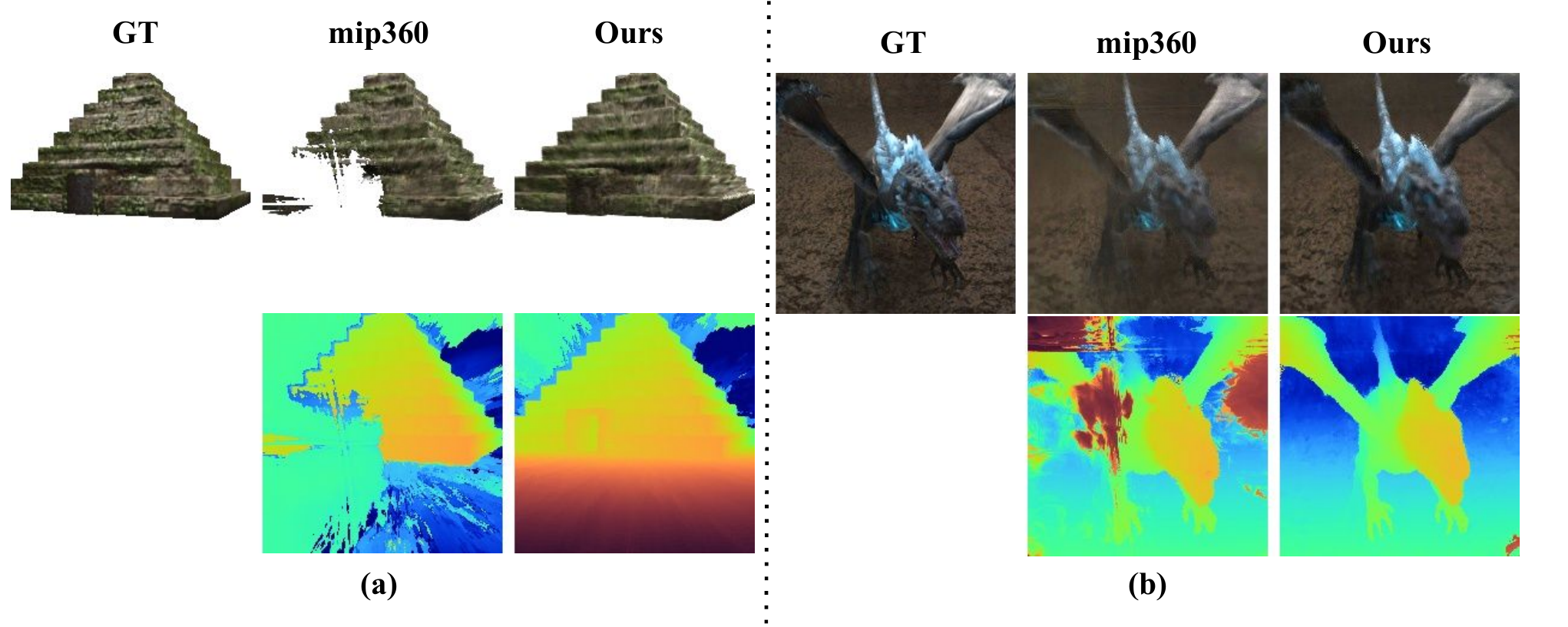}
    \caption{Qualitative Results for $2\times$ views on \textit{Dragon In Pyramid} scene. Observe that our results have less artefacts and much smoother depth maps.}
    \label{fig:qual_2x}
\end{figure}

\begin{figure}[!t]
    \centering
    \includegraphics[width=\linewidth]{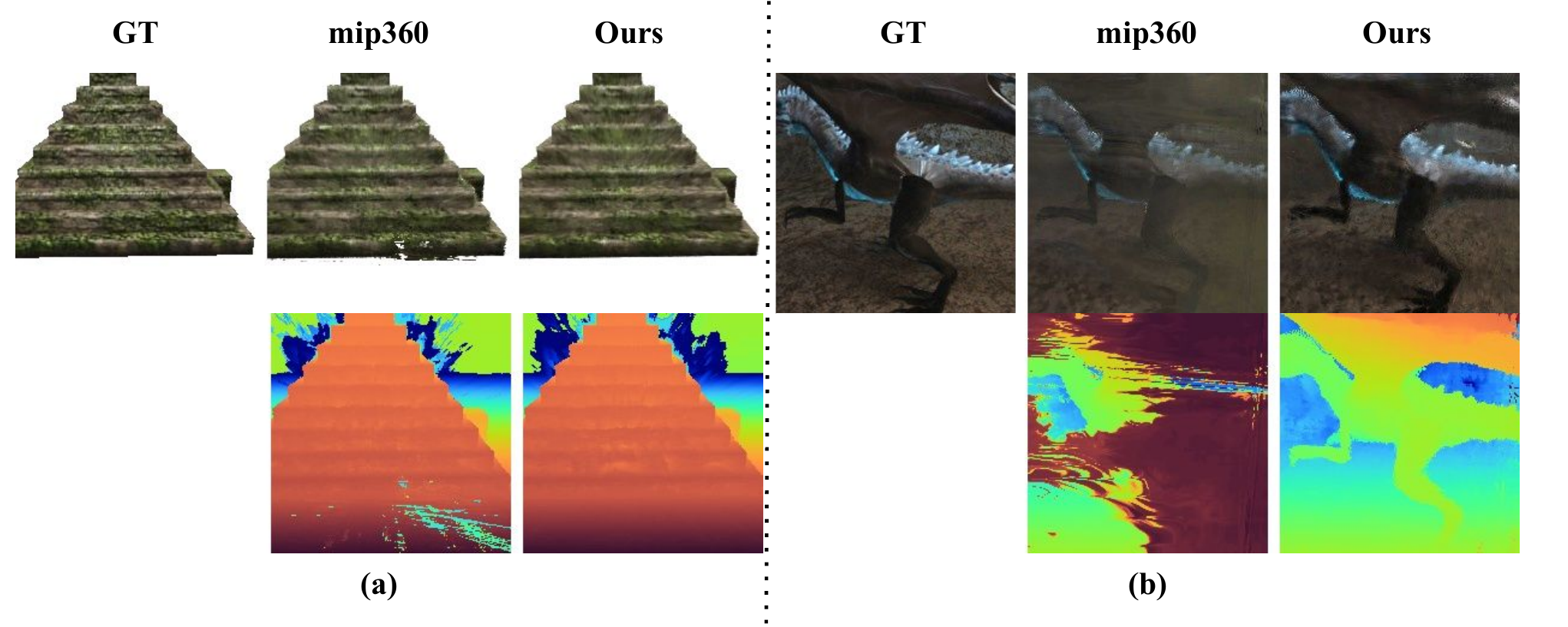}
    \caption{Qualitative Results for $3\times$ views on \textit{Dragon In Pyramid} scene. Observe that our results have less artefacts.}
    \label{fig:qual_3x}
\end{figure}

We present here another ablation which evaluates the effect of increasing number of views for a scene. Table \ref{tab:ablation_num_views_1} shows quantitative results on \textit{Dragon In Pyramid} scene by increasing number of views $2\times $ and $3\times$. Note that $2\times$ views mean that train, validation and test views will be doubled. We observe that as number of views are increased, overall metrics improves in both mip360~\cite{barron2022mip360} and our method. Further, we compare qualitative performance of our method with mip360~\cite{barron2022mip360} with increased number of views in Figure \ref{fig:qual_2x} and \ref{fig:qual_3x}. We observe that quality of depth map is much better in our method. Also, generated novel views from our method has less artefacts. 

\begin{table}[!t]
\caption{Performance on the \textit{Dragon In Pyramid} dataset for different number of views in the dataset. \textbf{Best} results are marked in bold. }
\label{tab:ablation_num_views_1}
\begin{adjustbox}{width=\linewidth}
\begin{tabular}{@{}cc|ccc|ccc|ccc@{}}
\toprule
\multicolumn{2}{c|}{\multirow{2}{*}{}} &
  \multicolumn{3}{c|}{\textbf{Level 0}} &
  \multicolumn{3}{c|}{\textbf{Level 1}} &
  \multicolumn{3}{c}{\textbf{Total}} \\ \cmidrule(l){3-11} 
\multicolumn{2}{c|}{} &
  \textbf{PSNR} &
  \textbf{SSIM} &
  \textbf{LPIPS} &
  \textbf{PSNR} &
  \textbf{SSIM} &
  \textbf{LPIPS} &
  \textbf{PSNR} &
  \textbf{SSIM} &
  \textbf{LPIPS} \\ \midrule
\multicolumn{1}{c|}{\multirow{2}{*}{\textbf{1x Views}}} & \textbf{mip360} & \textbf{30.8758} & \textbf{0.9006}   & \textbf{0.1367}   & {24.3890} &                    0.7054 &                    0.5163 & {27.6324} &                    0.8030 &                    0.3265  \\
\multicolumn{1}{c|}{}                                   & \textbf{Ours}  & 29.4773                    & 0.8700                    & 0.1699                    & \textbf{26.1722}   & \textbf{0.7489}   & \textbf{0.4573} & \textbf{27.8248}   & \textbf{0.8095} & \textbf{0.3136} \\ \midrule
\multicolumn{1}{c|}{\multirow{2}{*}{\textbf{2x Views}}} & \textbf{mip360} & \textbf{29.5127}   & \textbf{0.8436}   & \textbf{0.1830}   &{26.2172} &{0.7245} &{0.4627} &{27.8650} & {0.7841}   &{0.3228}  \\
\multicolumn{1}{c|}{}                                   & \textbf{Ours}   &{29.1104} &{0.8099} &{0.2176} & \textbf{27.4282}   & \textbf{0.7661}   & \textbf{0.4244}   & \textbf{28.2693}   & \textbf{0.7880} & \textbf{0.3210} \\ \midrule
\multicolumn{1}{c|}{\multirow{2}{*}{\textbf{3x Views}}} & \textbf{mip360}  & \textbf{31.1511}   & \textbf{0.8764}   & \textbf{0.1715}   & {26.5231} & {0.7239} & {0.4638} & {28.8371} & {0.8001}   & {0.3176}  \\
\multicolumn{1}{c|}{}                                   & \textbf{Ours}   & {30.5436} & {0.8461} & {0.1882} & \textbf{27.4354}   & \textbf{0.7693}   & \textbf{0.4385}   & \textbf{28.9895}   & \textbf{0.8077} & \textbf{0.3134}  \\ \bottomrule
\end{tabular}
\end{adjustbox}
\end{table}

\begin{figure*}[!t]
    \centering
    \includegraphics[width=\linewidth]{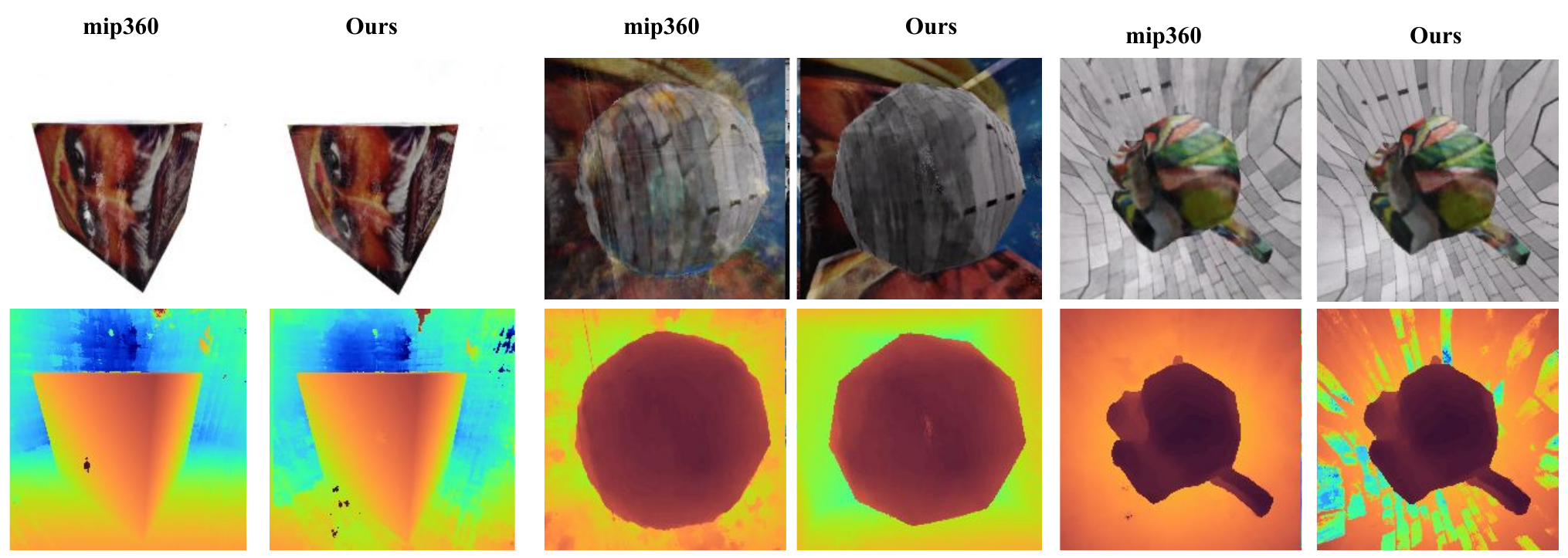}
    \caption{Qualitative comparison on OOD views. (\textbf{Top Row}) Generated novel views. (\textbf{Bottom Row}) Corresponding depth map. Check the quality of depth maps in inner levels for our method.  }
    \vspace{1em}
    \label{fig:ood_views}
\end{figure*}
\subsection{Out of Distribution Views}
The training set's views are uniformly sampled from the curved surface of a hemisphere with the camera's $z-axis$ always pointing towards the subject. Out-of-distribution (OOD) is any new view that does not lie on this hemisphere and whose $z-axis$ is not necessarily aligned with the subject. We investigated the quality of novel view synthesis for OOD views. We apply a random rotation and translation to the camera pose in the test set to produce OOD camera poses. A random translation value is sampled uniformly between $(10cm, 10cm)$, which is then used to translate the camera position along its $z-axis$. We randomly choose the rotation axis and angle from $(-45^{\circ}, 45^{\circ})$ for random rotation and change the current pose with this transformation. Figure \ref{fig:ood_views} shows the novel views and their corresponding depth maps. The depth map shows that our technique regularises the 3D geometry significantly better than other methods. Furthermore, the depth map quality is substantially better, which aids our method in producing non-blurry results.

\subsection{Additional Results}
\begin{figure*}[p]
    \centering
    \includegraphics[width=\linewidth]{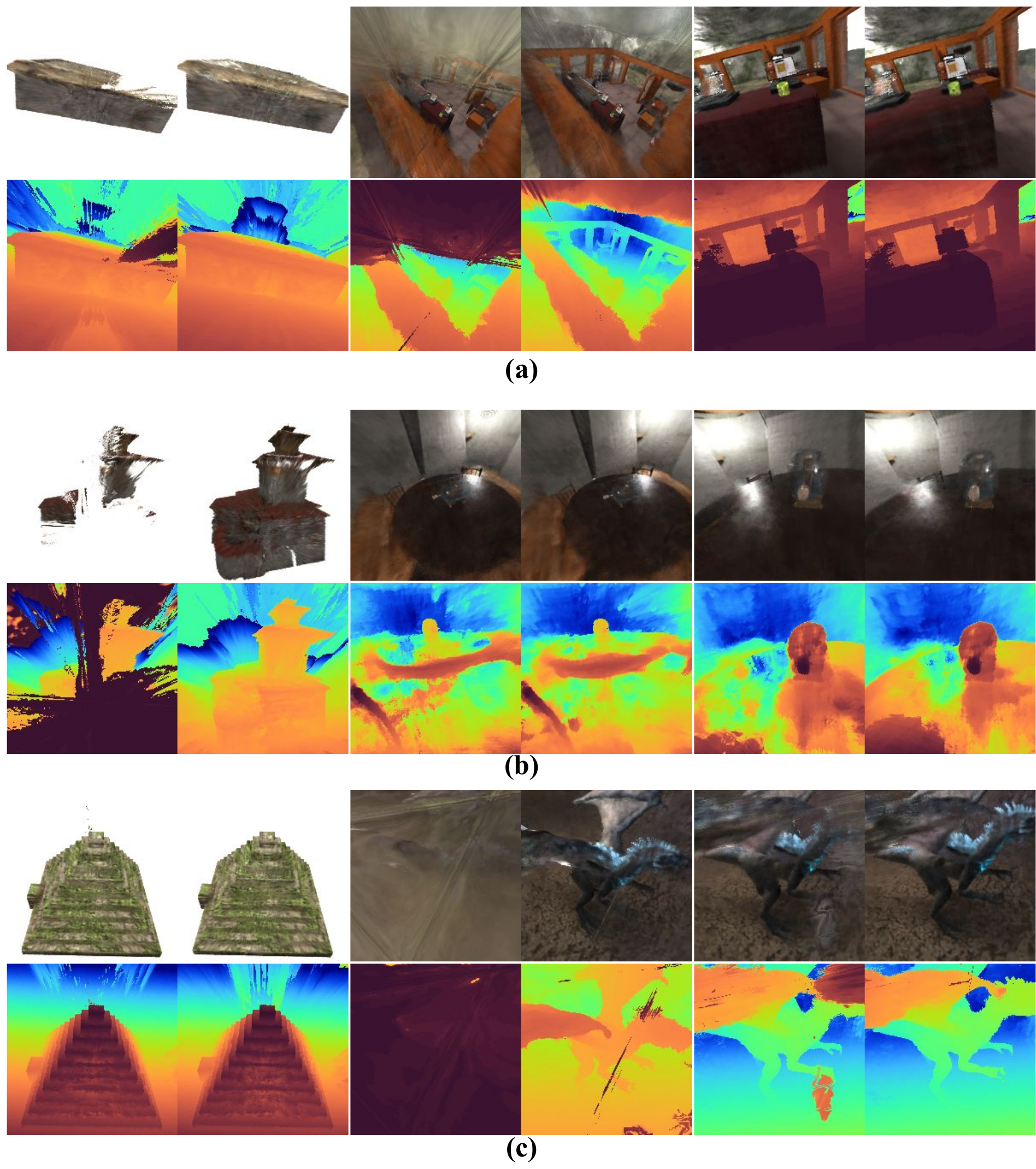}
    \caption{Out of distribution views for (a) Coffee Shop, (b) Bhutanese House and (c) Dragon In Pyramid Scene. \textbf{Odd} columns are results from Mip-NeRF 360~\cite{barron2022mip360} and \textbf{even} columns are results from our method. We observe that generated novel views from our method has less artefacts and better depth maps. Check the clarity in claws of dragon in last column of (c).}
    \label{fig:supp_ood}
\end{figure*}

\begin{figure*}[tp]
    \centering
    \includegraphics[width=\linewidth]{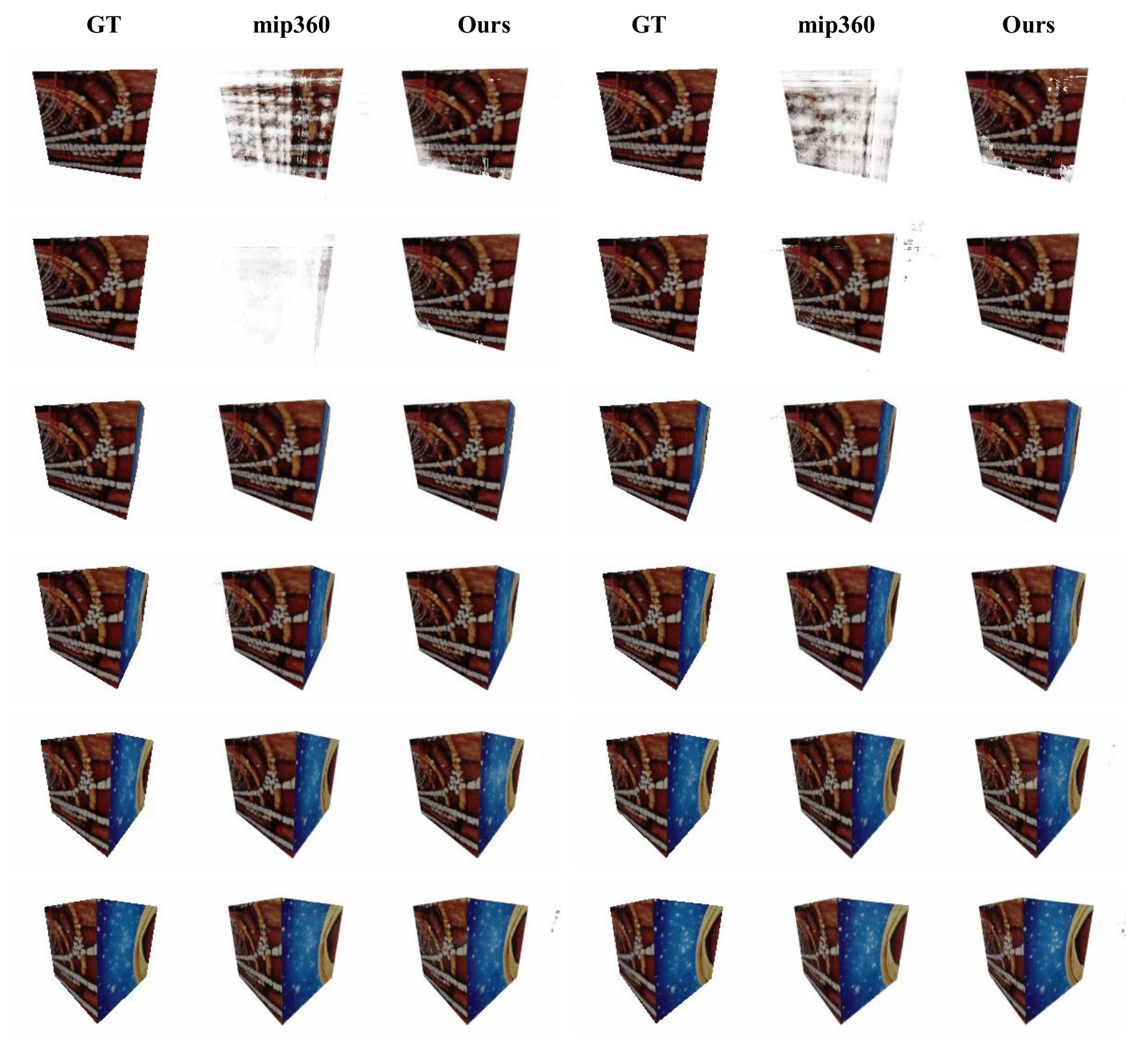}
    \caption{Sequence of generated novel views for Level 0 of \textit{Cube-Sphere-Monkey} scene. Please note that sequence is represented in zig-zag pattern. The generated novel views from our method has less artefacts. \textbf{Please check the video provided in the supplementary material to appreciate our results better.}}
    \label{fig:supp_unrolled_novel_views}
\end{figure*}

\begin{figure*}[tp]
    \centering
    \includegraphics[width=\linewidth]{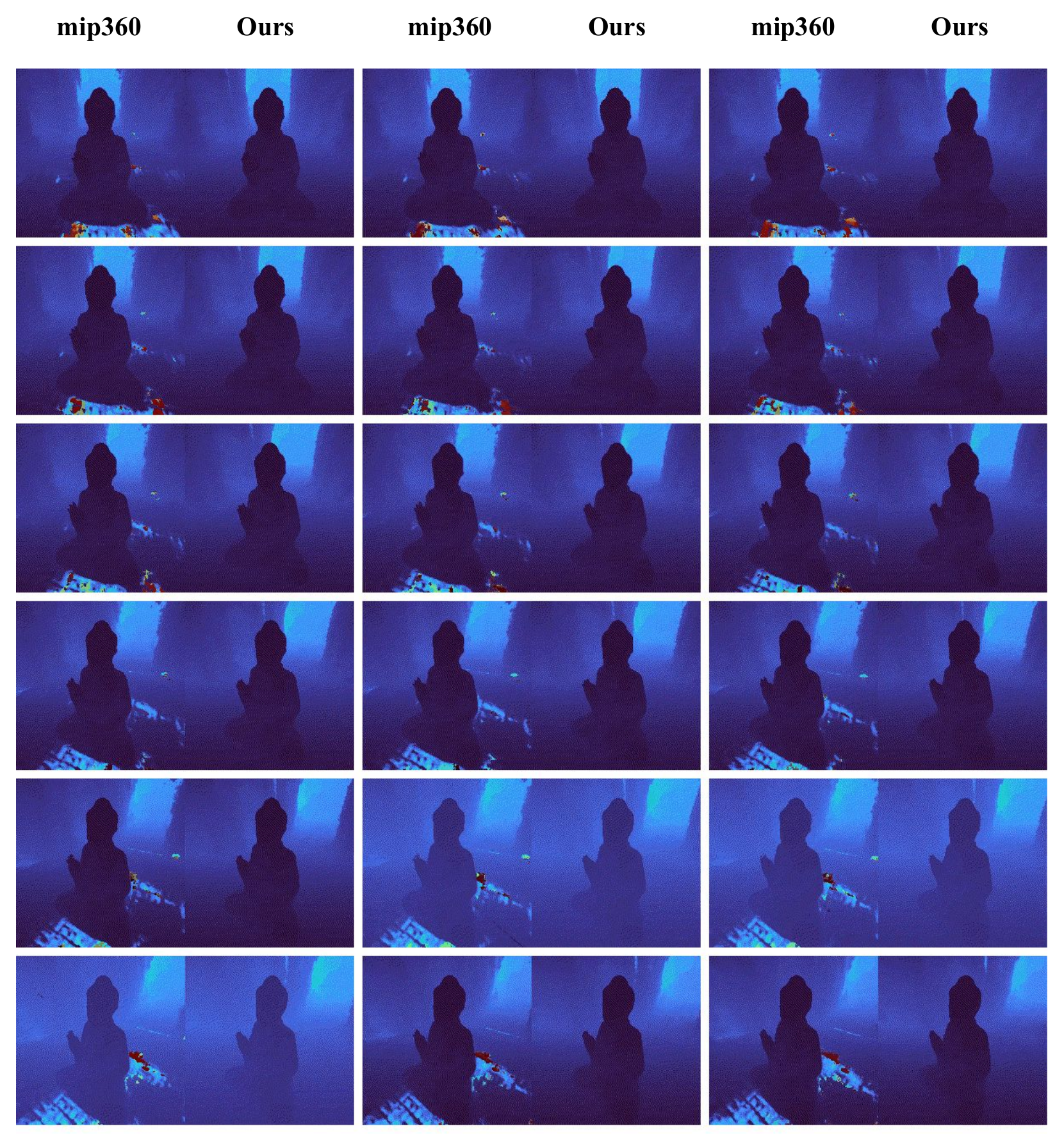}
    \caption{Sequence of depth maps of generated novel views for Level 1 of \textit{Buddhist Temple} scene. Please note that sequence is represented in zig-zag pattern. We observe that there is a collapse in the floor region for output from mip360~\cite{barron2022mip360} output. Whereas, our method generates smooth depth maps. \textbf{Please check the video provided in the supplementary material to appreciate our results better.}}
    \label{fig:supp_unrolled_heatmaps}
\end{figure*}

We provide more results for the Out Of Distribution views in Figure \ref{fig:supp_ood}. Further, we provide a sequence of generated novel views for \textit{Cube-Sphere-Monkey} in Figure \ref{fig:supp_unrolled_novel_views} and a sequence of depth maps for the  \textit{Buddhist Temple} in Figure \ref{fig:supp_unrolled_heatmaps}. There are distinct artefacts in column one and three in Figure \ref{fig:supp_ood}(a),  column one in \ref{fig:supp_ood}(b) and column three in \ref{fig:supp_ood}(b). We compare the generated depth maps in Figure \ref{fig:supp_ood} and Figure \ref{fig:supp_unrolled_heatmaps}. We observe that the depth maps from our method are smooth and have less artefacts than Mip-NeRF 360 ~\cite{barron2022mip360}. Notice the collapse in floor of the \textit{Buddhist Temple} scene in Figure \ref{fig:supp_unrolled_heatmaps}. From these results, it's clear that the generated novel views from our method has less artefacts and better 3D representation of such stratified scenes. 

\subsection{Impact of Image-Resolution on training.}
On $800 \times 800$ resolution for ``Cube-Sphere-Monkey'' scene, mipNeRF-360 achieves an overall PSNR of 23.17 and our method achieves 26.41. This is similar to behavior observed on low-resolution ($200 \times 200$)  and high-resolution RealEstate10K. 

\FloatBarrier
\onecolumn
\twocolumn

\newpage

{
\bibliographystyle{ieee_fullname}
\bibliography{arxiv}
}

\end{document}